\documentclass[lettersize,journal]{IEEEtran}
\usepackage{amsmath,amsfonts}
\usepackage[T1]{fontenc}
\usepackage{amssymb}
\usepackage{algorithmic}
\usepackage{algorithm}
\usepackage{array}
\usepackage[caption=false,font=normalsize,labelfont=sf,textfont=sf]{subfig}
\usepackage{textcomp}
\usepackage{stfloats}
\usepackage{url}
\usepackage{verbatim}
\usepackage{graphicx}
\usepackage{cite}
\usepackage{multirow}
\usepackage{makecell}

\begin{document}

\title{DTIF: Robust Loop Closure Detection via Delaunay Triangle Topology in Complex Forests}

\author{
\IEEEauthorblockN{
Xin Zhao,
Jianping Li, ~\IEEEmembership{Member,~IEEE},
Qin Zou, ~\IEEEmembership{Senior Member,~IEEE},\\
Fuxun Liang, 
Zhen Dong, ~\IEEEmembership{Senior Member,~IEEE},
Bisheng Yang}
\thanks{
This study was jointly supported by the National Key Research and Development Program of China (No. 2024YFF1308300) and Wuhan Natural Science Foundation Project (No. 2025041001010363). (Corresponding author: Qin Zou)

Xin Zhao and Qin Zou are with the School of Computer Science, Wuhan University, Wuhan 430079, China (e-mails: xinzhaodc@whu.edu.cn, qzou@whu.edu.cn).

Jianping Li is with the School of Electrical and Electronic Engineering, Nanyang Technological University, 50 Nanyang Avenue 639798, Singapore (e-mail: jianping.li@ntu.edu.sg).

Fuxun Liang is with the School of Urban Design, Wuhan University, Wuhan 430079, China (e-mail: liangfuxun@whu.edu.cn).

Zhen Dong and Bisheng Yang are with the LIESMARS, Wuhan University, Wuhan 430079, China (e-mails: dongzhenwhu@whu.edu.cn, bshyang@whu.edu.cn).

}
}


\maketitle

\begin{center}
\footnotesize
This work has been submitted to the IEEE for possible publication.
Copyright may be transferred without notice, after which this version
may no longer be accessible.
\end{center}

\begin{abstract}
Accurate forest inventory and large-scale mapping are essential for ecosystem monitoring and sustainable forest management. Multiple low-cost edge platforms enable efficient large-area data acquisition, but merging independently constructed local maps in GNSS-denied understory environments still requires initialization-free loop closure detection and global registration. This task is challenging because low-cost LiDAR point clouds are sparse and noisy, while repetitive trunk layouts and the lack of distinctive geometric landmarks lead to severe perceptual aliasing and false correspondences. To address these issues, we propose DTIF (Delaunay Triangulation in Forests), a lightweight trunk-topology-based framework for forest loop closure detection and global registration. Tree trunks are first extracted as stable landmarks and encoded using a Delaunay topology for compact scene representation. Candidate submaps are then screened using edge-length and radius statistics, followed by edge--radius consistency verification and strong/weak vertex support aggregation to construct weighted vertex correspondences. Finally, topology-derived reliability weights are incorporated into a decoupled robust pose estimator that separately estimates yaw, horizontal translation, and elevation translation under gravity alignment. Experiments on simulated and real-world forest datasets demonstrate that DTIF achieves accurate registration with low computational overhead, providing a favorable balance among robustness, efficiency, and deployability on resource-constrained edge platforms.
\end{abstract}


\section{Introduction} \label{section:1}

\IEEEPARstart{F}{orests} play an indispensable role in maintaining ecological balance, regulating the global carbon cycle, and preserving biodiversity \cite{mitchardTropicalForestCarbon2018}. Accurate estimation of forest structural parameters supports ecosystem monitoring, environmental change assessment, and forest dynamics analysis \cite{wangReviewRemoteSensing2019, xiaoRemoteSensingTerrestrial2019}, while providing a basis for sustainable forest management and resource planning \cite{liuEstimatingForestStructural2018, nayhaTransitionFinnishForestbased2019}. As forest inventory expands toward larger spatial scales, higher revisit frequencies, and finer levels of detail, conventional approaches such as manual measurement and terrestrial laser scanning (TLS) are increasingly constrained by high labor costs and limited operational efficiency \cite{zengNationalForestInventory2015, liangTerrestrialLaserScanning2016}. Mobile mapping platforms, including unmanned aerial vehicles (UAVs), unmanned ground vehicles (UGVs), and wearable edge devices, have therefore been increasingly adopted for forest resource surveys \cite{liangInsituMeasurementsMobile2018, liuEstimatingForestStructural2018a, liWHUHelmetHelmetbasedMultisensor2023, chudaHANDHELDMOBILELASER2020, mokrosNovelLowcostMobile2021}. Among them, low-cost edge platforms are particularly attractive because of their portability, real-time processing capability, and deployment flexibility \cite{suDevelopmentEvaluationBackpack2021, proudmanRealtimeForestInventory2022, liRealtimeAutomatedForest2023}. In large-area forest surveys, multiple operators often acquire local maps independently. Since understory environments are typically GNSS-denied, these maps lack a common global reference and cannot be directly integrated \cite{caboComparingTerrestrialLaser2018}. Reliable loop closure detection and global registration without prior pose information are therefore essential for merging independently acquired local maps into a unified forest representation.

This task remains highly challenging in forest understory environments. Due to payload and cost constraints, edge platforms are commonly equipped with low-cost LiDAR sensors that produce sparse observations affected by measurement noise and environmental clutter \cite{mitkaProcedureAccuracyAssessment2025}. Unlike urban and indoor scenes with abundant planar surfaces and distinctive landmarks, natural forests generally lack stable and highly discriminative geometric structures \cite{dongRegistrationLargescaleTerrestrial2020a, tremblayAutomatic3DMapping2019}. Repetitive trunk arrangements can produce similar local layouts at different locations, leading to severe perceptual aliasing. This structural repetition increases false correspondences and aggravates pose-estimation degeneracy, thereby reducing the reliability of loop closure detection and global registration.

Existing place recognition and global registration approaches generally comprise two stages: front-end correspondence construction and back-end pose estimation. Front-end methods mainly rely on handcrafted descriptors \cite{rusuFastPointFeature2009, kimScanContextEgocentric2018}, topological graph matching \cite{yuanSTDStableTriangle2023, yuanBTCBinaryTriangle2024}, or deep learning models \cite{uyPointNetVLADDeepPoint2018}, whereas back-end methods recover relative poses through random sampling, robust optimization, or certifiable registration. Despite substantial progress, two limitations remain. First, most correspondence construction methods focus on local geometric features or learned global representations while insufficiently exploiting the stable spatial topology of tree trunks, one of the few repeatable geometric cues in forest environments. Consequently, they may generate numerous redundant or ambiguous candidate correspondences, increasing computational cost and reducing matching reliability. Second, many robust registration frameworks solve a unified six-degree-of-freedom problem \cite{yangTEASERFastCertifiable2021} without explicitly considering the different constraint characteristics of horizontal and vertical directions in forest scenes. In addition, correspondence reliability is rarely propagated directly into the pose optimization \cite{yanNewOutlierRemoval2022}, limiting robustness and efficiency under high-outlier conditions.

To overcome these limitations, we propose a lightweight and robust framework for loop closure detection and global registration in forest understory environments. The core idea is to replace raw point-level feature matching with trunk-topology-guided correspondence construction and reliability-aware pose estimation. \textbf{A trunk-topology representation} uses tree trunks as repeatable structural landmarks and constructs a compact Delaunay topology from their horizontal centers, reducing dependence on local point features affected by LiDAR noise, foliage, shrubs, and occlusion. \textbf{A strong and weak vertex-pair support mechanism} verifies canonical triangle attributes through edge-length and endpoint-radius consistency and accumulates repeated local evidence into vertex-pair support statistics, producing weighted vertex correspondences. \textbf{A reliability-weighted decoupled robust pose estimator} propagates these topology-derived weights into back-end registration and decomposes pose estimation into yaw rotation, horizontal translation, and elevation translation under the gravity-aligned assumption. Together, these designs improve correspondence reliability, registration robustness, and computational efficiency for large-scale forest map fusion on resource-constrained edge platforms.

The remainder of this paper is organized as follows. Section~\ref{section:2} reviews the related work. Section~\ref{section:3} presents the proposed loop closure detection and robust decoupled pose estimation methodology. Section~\ref{section:4} reports and discusses the experimental results on simulated and real-world datasets. Section~\ref{section:5} concludes the paper and outlines future research directions.

\section{Related Work} \label{section:2} 

Achieving robust place recognition and global registration from 3D point clouds remains a challenging problem in unstructured environments, particularly in forest understory scenes. Due to the sparse and noisy observations acquired by low-cost LiDAR sensors, existing methods often struggle to simultaneously achieve high robustness and computational efficiency. This section reviews related studies from two perspectives: loop closure descriptors and robust global registration methods.

\subsection{Loop Closure Descriptors}
\textbf{Handcrafted Global Descriptors}: Early research on point cloud registration and loop closure detection primarily focused on extracting geometric features or statistical properties from sensor observations to construct robust scene representations. Traditional local descriptors, such as Fast Point Feature Histograms (FPFH) \cite{rusuFastPointFeature2009}, SHOT \cite{saltiSHOTUniqueSignatures2014}, and Spin Images \cite{johnsonUsingSpinImages1999}, encode the relationships among surface normals and local curvatures within a neighborhood to obtain rotation- and translation-invariant representations. These approaches have demonstrated strong performance in object-level and fine-grained registration tasks. As large-scale mapping applications emerged, a variety of global descriptors were developed. M2DP \cite{heM2DPNovel3D2016a} projects 3D point clouds onto multiple 2D planes to generate density distributions and subsequently derives a compact global representation through singular value decomposition. Similarly, the widely used Scan Context family, including Scan Context \cite{kimScanContextEgocentric2018} and Scan Context++ \cite{kimScanContextStructural2022}, inspired by \cite{belongieShapeMatchingObject2002a}, projects point clouds into polar or Cartesian coordinate systems and constructs global feature maps encoding height distribution information for efficient place retrieval. Despite their effectiveness in structured environments such as urban roads and indoor scenes, these handcrafted descriptors rely heavily on distinctive geometric structures, including building facades, corners, and planar surfaces. In forest environments, however, the lack of discriminative geometric landmarks, together with severe environmental clutter and measurement noise, often leads to ambiguous descriptor representations and a large number of false correspondences, thereby degrading loop closure detection performance.

\textbf{Topological and Graph-Based Matching}: To improve robustness in unstructured environments, numerous studies have incorporated graph and topological structures into the correspondence construction process. RTAB-Map \cite{labbeAppearanceBasedLoopClosure2013} performs loop closure detection by representing frequently revisited locations as graph landmarks. In forest environments, \cite{nardariPlaceRecognitionForests2021} proposed a descriptor based on the Urquhart graph, utilizing tree locations to establish correspondences between observations and previously visited landmarks. Similarly, \cite{jiangTriangleFeatureBased2019a} constructed two-dimensional triangular structures from corner features for loop closure detection in 2D SLAM systems. Building upon this idea, \cite{yuanSTDStableTriangle2023} further extracted three-dimensional triangular primitives and performed keyframe registration according to the matching criteria proposed in \cite{umeyamaLeastsquaresEstimationTransformation1991}, while \cite{yuanBTCBinaryTriangle2024} introduced additional binary environmental descriptors to improve discriminative capability. Although topological representations generally provide stronger robustness than purely descriptor-based approaches, most existing methods construct topologies from large numbers of local geometric primitives. As scene scale increases, these representations tend to generate substantial redundancy, resulting in a rapidly growing number of candidate correspondences. In forest environments, where repetitive structures are common, such redundancy often introduces numerous ineffective geometric primitives, negatively affecting both matching efficiency and registration accuracy.

\textbf{Learning-Based Place Recognition}: In recent years, deep learning has become an important research direction for point cloud place recognition. PointNetVLAD \cite{uyPointNetVLADDeepPoint2018} combines PointNet \cite{charlesPointNetDeepLearning2017} and NetVLAD \cite{arandjelovicNetVLADCNNArchitecture2018a} within an end-to-end learning framework to generate discriminative global descriptors, significantly improving place recognition performance in large-scale environments. Subsequent methods, such as LPD-Net \cite{liuLPDNet3DPoint2019} and MinkLoc3D \cite{komorowskiMinkLoc3DPointCloud2021}, further enhance representation capability by incorporating local geometric structures and sparse convolutional operations. More recently, Transformer-based architectures \cite{vaswaniAttentionAllYou2017} have been introduced into place recognition tasks. Methods such as TransLoc3D \cite{xuTransLoc3DPointCloud2023} leverage long-range feature interactions to obtain more expressive scene representations. Beyond directly processing raw point clouds, several studies have adopted bird’s-eye-view (BEV) representations for place recognition, including Ring++ \cite{xuRINGRotoTranslationInvariant2023} and BEVPlace++ \cite{luoBEVPlaceFastRobust2025}. For forest environments specifically, ForestLPR \cite{shenForestLPRLiDARPlace2025} generates multi-layer BEV density maps from different height slices and learns complementary information from tree trunks, shrubs, and other vegetation layers to alleviate the severe self-similarity problem commonly observed in forests. Despite their impressive performance, learning-based methods typically require large-scale training datasets and GPU acceleration. Their convolutional, Transformer, or visual backbone networks often incur considerable computational and memory overhead, making deployment on low-cost resource-constrained edge platforms challenging.

In summary, significant progress has been made in loop closure detection and place recognition for structured environments. Handcrafted descriptors offer low computational complexity but are highly sensitive to structural degradation and environmental noise in forest scenes. Topological and graph-based methods improve robustness by exploiting structural relationships; however, they often rely on large numbers of local geometric primitives, resulting in reduced efficiency as scene scale increases. Learning-based approaches provide stronger representation capabilities but typically require extensive training data and substantial computational resources. Therefore, developing an efficient correspondence construction framework that can effectively exploit the stable and readily available trunk topology in forest environments while maintaining low computational cost remains an important open challenge.

\subsection{Robust Global Registration}
\textbf{Correspondence Sampling}: After obtaining candidate correspondences, accurately estimating the relative pose between two point clouds in the presence of large numbers of mismatches remains a fundamental challenge in global registration. Early approaches mainly relied on Random Sample Consensus (RANSAC) \cite{fischlerRandomSampleConsensus1981} and its variants, which repeatedly sample minimal constraint sets and verify geometric consistency to estimate optimal transformation parameters. SAC-IA \cite{weiSACIAAlgorithmBased2023} further improves sampling efficiency by incorporating local geometric features, while 4PCS \cite{aiger4pointsCongruentSets} and Super4PCS \cite{melladoSuper4PCSFast2014} exploit affine-invariant geometric primitives to perform coarse registration without requiring an initial pose estimate. Nevertheless, these methods fundamentally depend on random sampling. As the outlier ratio increases, the probability of selecting a valid minimal set decreases rapidly, resulting in substantially increased computational complexity.

\textbf{Robust and Certifiable Registration}: To improve registration efficiency and robustness, researchers have increasingly incorporated robust optimization techniques into global registration frameworks. Fast Global Registration (FGR) \cite{zhouFastGlobalRegistration2016} jointly suppresses incorrect correspondences and estimates poses within a unified optimization framework, enabling efficient initialization-free registration. Subsequently, methods based on Graduated Non-Convexity (GNC) \cite{yangGraduatedNonConvexityRobust2020a} and robust M-estimators have been widely adopted to further improve tolerance against outliers while maintaining computational efficiency. As registration scenarios become increasingly challenging, research attention has gradually shifted from optimization efficiency toward robustness under extreme outlier conditions. One line of work constructs consistency graphs from pairwise geometric relationships and identifies highly reliable correspondence subsets. Representative examples include PMC \cite{parraPracticalMaximumClique} and CLIPPER \cite{luskCLIPPERGraphTheoreticFramework2021}, which utilize maximum clique search or consistency graph optimization to identify globally consistent correspondences. GROR \cite{yanNewOutlierRemoval2022} further exploits node and edge reliability within correspondence graphs to progressively refine high-quality inlier sets, enabling efficient registration under high-outlier conditions. Another line of research focuses on certifiable registration frameworks with theoretical guarantees. GORE \cite{parrabustosGuaranteedOutlierRemoval2018} progressively eliminates correspondences that cannot belong to the optimal consensus set through deterministic pruning strategies, thereby significantly reducing the search space. TEASER++ \cite{yangTEASERFastCertifiable2021} decomposes the registration problem into scale, rotation, and translation estimation using invariant measurements and combines maximum clique search with truncated least-squares optimization to achieve certifiable global registration. Compared with traditional sampling-based approaches, these methods maintain stable performance even under extremely high outlier ratios.

\textbf{Degeneracy-Aware Registration}: Beyond outlier rejection, recent studies have shown that pose estimation may still become degenerate under low-overlap conditions, large viewpoint changes, or long-range loop closures, even when incorrect correspondences are successfully removed. To address this issue, Quatro \cite{limSingleCorrespondenceEnough2022} introduces environmental structural priors to constrain rotation estimation and decouples the pose estimation process, thereby reducing the minimum number of correspondences required for successful registration. Quatro++ \cite{limQuatroRobustGlobal2024} further incorporates additional scene priors, such as ground segmentation, to improve robustness and registration success rates in loop closure and global localization tasks. These studies suggest that exploiting meaningful scene priors and reducing optimization degrees of freedom can effectively alleviate degeneracy under challenging registration conditions.

However, most existing robust global registration methods are primarily designed for structured environments such as urban roads and indoor scenes, where stable planar structures are readily available. In forest environments, severe vegetation occlusion and highly repetitive local geometries often lead to correspondence degeneration and unstable pose estimation. Moreover, some methods require constructing large-scale consistency graphs and performing computationally expensive searches, whereas others depend heavily on environment-specific assumptions. These limitations restrict their practical deployment on low-cost edge platforms. Therefore, developing a computationally efficient registration framework that can effectively exploit stable topological information in forest environments while maintaining robustness under high-outlier conditions remains an open research problem.

\section{Methodology}\label{section:3} 

\subsection{Problem Formulation and System Overview}

\begin{figure*}[!t]
\centering
\includegraphics[width=0.95\textwidth]{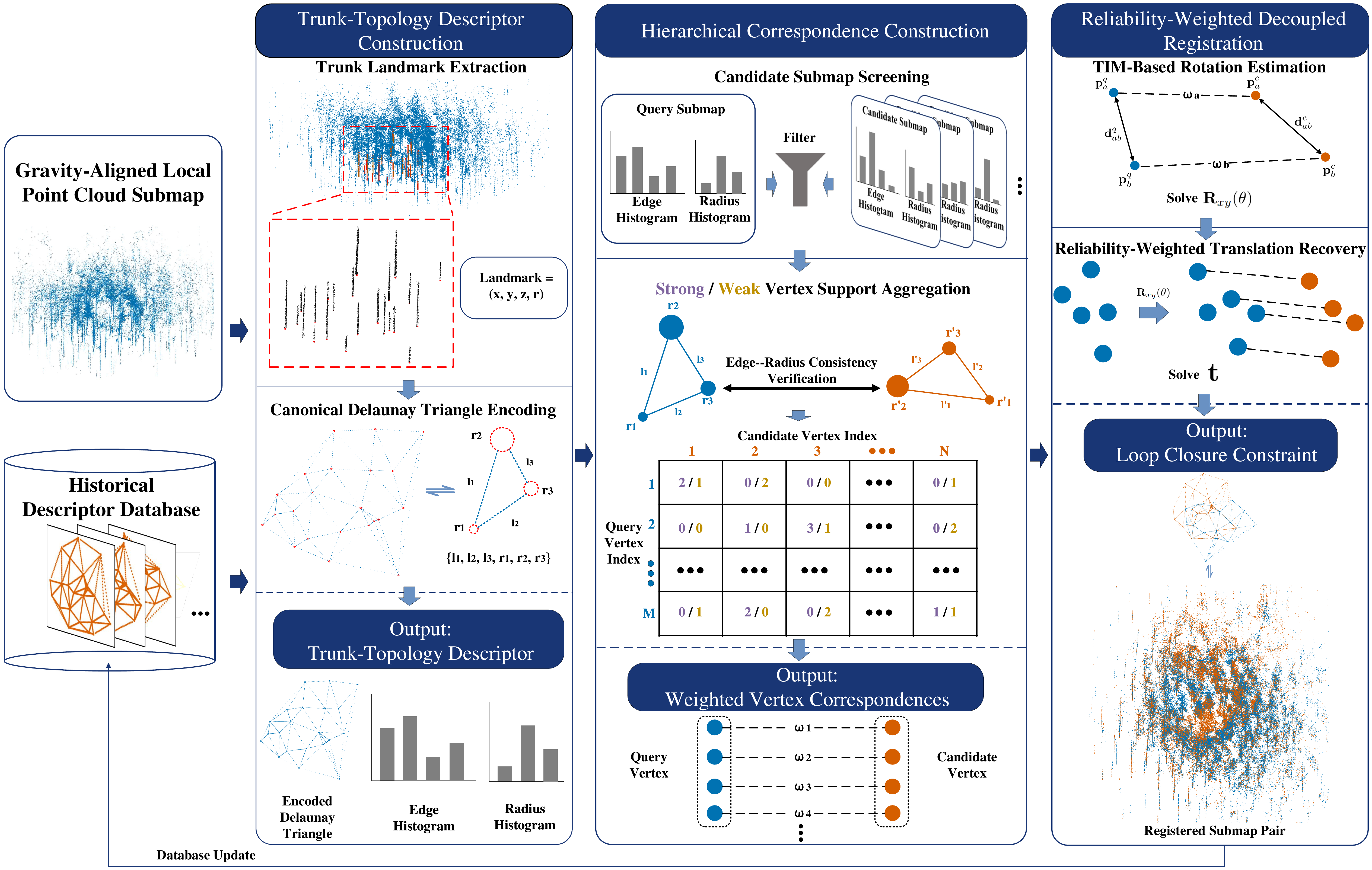}
\caption{Overview of the proposed trunk-topology-based loop closure detection and global registration framework.}
\label{fig_1}
\end{figure*}

This study focuses on loop closure detection and relative pose estimation between local forest submaps in GNSS-denied understory environments. The input submaps are constructed and gravity-aligned by Fast-LIO2 \cite{xuFASTLIO2FastDirect2022}. Given a query submap, the goal is to retrieve overlapping historical candidate submaps and estimate their relative transformations for subsequent multi-submap fusion.

As shown in Fig.~\ref{fig_1}, the proposed framework consists of three main modules. The first module constructs a compact trunk-topology descriptor through trunk landmark extraction and canonical Delaunay triangle encoding, yielding the Delaunay topology together with edge and radius histograms. The second module performs hierarchical correspondence construction, where candidate submaps are screened and local edge--radius consistency is accumulated as strong and weak vertex support to generate weighted vertex correspondences. The third module estimates the relative transformation through reliability-weighted decoupled registration, including TIM-based horizontal rotation estimation and reliability-weighted translation estimation.

For the $i$-th submap, let $\mathcal{V}_i$ and $\mathcal{T}_i$ denote the extracted trunk landmarks and the Delaunay topology, respectively. The submap descriptor is denoted as $\mathcal{D}_i=(\mathcal{V}_i,\mathcal{T}_i,\mathbf{h}_i^e,\mathbf{h}_i^r)$, where $\mathbf{h}_i^e$ is the edge-length histogram constructed from the unique edges of the Delaunay triangulation, and $\mathbf{h}_i^r$ is the trunk-radius histogram. For a query descriptor $\mathcal{D}_q$ and a candidate descriptor $\mathcal{D}_c$, the front-end module produces a weighted vertex correspondence set $\mathcal{C}_{qc}=\{(\mathbf{p}_k^q,\mathbf{p}_k^c,\omega_k)\}_{k=1}^{K}$, where $\mathbf{p}_k^q$ and $\mathbf{p}_k^c$ are the matched 3-D vertex positions in the query and candidate submaps, respectively, and $\omega_k$ is the topology-derived reliability weight.

Under the gravity-aligned assumption, the relative transformation from the query submap to the candidate submap is modeled as
\begin{equation}
\label{Eq_3.1}
\mathbf{p}_k^c
=
\mathbf{R}_z(\theta)\mathbf{p}_k^q
+
\mathbf{t}_{qc}
+
\boldsymbol{\epsilon}_k,
\quad
\mathbf{t}_{qc}=(t_x,t_y,t_z)^\top ,
\end{equation}
where $\mathbf{R}_z(\theta)$ denotes the yaw rotation around the gravity direction, $\mathbf{t}_{qc}$ is the relative translation, and $\boldsymbol{\epsilon}_k$ is the residual term. The estimated transformation $(\mathbf{R}_{qc},\mathbf{t}_{qc})$ is used as a loop closure constraint for global forest map construction.

For sequential loop closure processing, each valid query descriptor is matched against the historical descriptor database $\mathcal{B}$ before being inserted into the database. Each accepted match outputs a loop closure constraint $(q,c,\mathbf{R}_{qc},\mathbf{t}_{qc})$ for subsequent multi-submap fusion.

\subsection{Lightweight Scene Representation Based on Stable Trunk Topology}

Each gravity-aligned forest submap is converted into a compact trunk-topology descriptor.

\noindent\textbf{Trunk landmark extraction. }Following the vertical-continuity-based trunk extraction idea in \cite{zhangTerrainIndividualTree2025}, trunk landmarks are extracted without explicit ground segmentation. High-density voxel columns with sufficient vertical continuity and plausible lower elevations are retained and merged into trunk candidates. After outlier removal, circle fitting is performed to estimate the trunk center and radius. Each valid trunk landmark is represented as $\mathbf{v}_m=(x_m,y_m,z_m,r_m)$, where $(x_m,y_m)$ is the horizontal center, $z_m$ is the estimated base elevation, and $r_m$ is the radius attribute.

\noindent\textbf{Canonical Delaunay triangle encoding. }Given the valid trunk landmarks $\mathcal{V}=\{\mathbf{v}_m\}_{m=1}^{M}$, a two-dimensional Delaunay triangulation is constructed on their horizontal centers $\{(x_m,y_m)\}_{m=1}^{M}$. Each finite non-degenerate triangular face is retained as a local topological primitive, and the retained primitives form the Delaunay topology $\mathcal{T}$. Submaps with too few valid trunks or triangular primitives are excluded from loop closure retrieval. The unique edges of $\mathcal{T}$ are used to construct the edge-length histogram $\mathbf{h}^e$, while the radii of valid trunks are used to construct the radius histogram $\mathbf{h}^r$.

For each retained triangle $\tau$ with vertices $(u,v,w)$, we define its edge--length pair set as
$\mathcal{E}_{\tau}=\{(e_{uv},\ell_{uv}),(e_{vw},\ell_{vw}),(e_{wu},\ell_{wu})\}$,
where $e_{uv}$ is a topological edge and $\ell_{uv}$ is its corresponding horizontal length. 
To remove vertex-order ambiguity, these edge--length pairs are sorted by decreasing length:
\begin{equation}
\label{Eq_3.2}
\begin{aligned}
\bigl((e_a,\ell_a)\bigr)_{a=1}^{3}
&=
\operatorname{sort}_{\downarrow}^{\ell}
(\mathcal{E}_{\tau}),\\
\boldsymbol{\ell}_{\tau}
&=
(\ell_1,\ell_2,\ell_3),
\quad
\ell_1\geq \ell_2\geq \ell_3,\\
\mathcal{I}_{\tau}
&=
(e_1\cap e_2,\;e_2\cap e_3,\;e_3\cap e_1).
\end{aligned}
\end{equation}
Here, $(e_a,\ell_a)$ is the $a$-th ordered edge--length pair, and $e_a\cap e_b$ denotes the vertex shared by two ordered edges. The tuple $\mathcal{I}_{\tau}=(i_1,i_2,i_3)$ defines the canonical vertex order of triangle $\tau$.

The vertex radii are rearranged according to this canonical order, and the triangle attribute is encoded as
\begin{equation}
\label{Eq_3.3}
\begin{aligned}
\mathbf{r}_{\tau}
&=
(r_{i_1},r_{i_2},r_{i_3}),\\
\tau
&=
(\mathcal{I}_{\tau},\boldsymbol{\ell}_{\tau},\mathbf{r}_{\tau}).
\end{aligned}
\end{equation}
Thus, each triangle is represented by ordered edge lengths and order-consistent vertex radii. These canonical attributes are used as the basic structural units for edge--radius consistency verification and vertex-support aggregation.

\subsection{Hierarchical Screening with Vertex Support Aggregation}

Given a query descriptor $\mathcal{D}_q$ and the historical descriptor database $\mathcal{B}$, this module retrieves candidate descriptors and outputs weighted vertex correspondences $\mathcal{C}_{qc}$, as summarized in Algorithm~\ref{alg:alg1}.

\noindent\textbf{Candidate submap screening. }For each submap, two normalized histograms are maintained: the edge-length histogram $\mathbf{h}^e$ computed from the unique edges of the Delaunay triangulation, and the trunk-radius histogram $\mathbf{h}^r$ computed from valid trunk radii. For a query descriptor $\mathcal{D}_q$ and a historical descriptor $\mathcal{D}_c$, their histogram similarities are measured by histogram intersection. A historical descriptor is retained only when both the edge-length and radius distributions are sufficiently similar to those of the query descriptor. This coarse screening step removes obviously unrelated submaps and passes the retained candidates to triangle-level consistency verification.

\noindent\textbf{Edge–-radius verification with vertex support aggregation. }For each retained query--candidate descriptor pair, Delaunay triangles are compared according to their canonical attributes. For a triangle pair $(\tau_q,\tau_c)\in\mathcal{T}_q\times\mathcal{T}_c$, the ordered edge-length vectors are $\boldsymbol{\ell}_{\tau_q}=(\ell_1^q,\ell_2^q,\ell_3^q)$ and $\boldsymbol{\ell}_{\tau_c}=(\ell_1^c,\ell_2^c,\ell_3^c)$, and the order-consistent radius vectors are $\mathbf{r}_{\tau_q}=(r_1^q,r_2^q,r_3^q)$ and $\mathbf{r}_{\tau_c}=(r_1^c,r_2^c,r_3^c)$. Instead of assigning a single similarity score to the whole triangle pair, each canonical edge is verified together with the radii of its two endpoint vertices.

The endpoint positions of the $a$-th canonical edge are denoted by $\partial e_a=\{u_a,v_a\}$. The edge--radius consistency indicator is defined as
\begin{equation}
\label{Eq_3.4}
\begin{aligned}
\chi_a(\tau_q,\tau_c)
=
\mathbb{I}\Big[
&\operatorname{Cons}_{\ell}(\ell_a^q,\ell_a^c)
\ \land \\
&\operatorname{Cons}_{r}(r_{u_a}^q,r_{u_a}^c)
\ \land \\
&\operatorname{Cons}_{r}(r_{v_a}^q,r_{v_a}^c)
\Big],
\quad a=1,2,3,
\end{aligned}
\end{equation}
where $\operatorname{Cons}_{\ell}(\cdot,\cdot)$ and $\operatorname{Cons}_{r}(\cdot,\cdot)$ indicate whether two edge lengths or two radii are consistent under predefined relative discrepancy tolerances. The verified edge set is written as
$\mathcal{A}_{\tau_q,\tau_c}=\{a\mid \chi_a(\tau_q,\tau_c)=1\}$.

The verified edge set provides local structural evidence for vertex-pair support aggregation. For each retained query--candidate descriptor pair, two vertex-pair support matrices are initialized as
\begin{equation}
\label{Eq_3.5}
\mathbf{E}^{s},\mathbf{E}^{w}
\in
\mathbb{N}^{|\mathcal{V}_q|\times|\mathcal{V}_c|},
\quad
\mathbf{E}^{s}=\mathbf{0},
\quad
\mathbf{E}^{w}=\mathbf{0}.
\end{equation}
The entries $E^s_{mn}$ and $E^w_{mn}$ record the strong and weak support counts for the vertex pair formed by the $m$-th query trunk and the $n$-th candidate trunk, respectively.

For a triangle pair $(\tau_q,\tau_c)$, the canonical vertex orders are written as
$\mathcal{I}_{\tau_q}=(i_1^q,i_2^q,i_3^q)$ and
$\mathcal{I}_{\tau_c}=(i_1^c,i_2^c,i_3^c)$. 
If all three canonical edges are verified, the triangle pair provides complete triangle-level evidence and contributes strong support to the three canonical vertex pairs:
\begin{equation}
\label{Eq_3.6}
E^s_{i_b^q i_b^c}
\leftarrow
E^s_{i_b^q i_b^c}+1,
\quad
b=1,2,3,
\quad
\mathrm{if}\ 
|\mathcal{A}_{\tau_q,\tau_c}|=3.
\end{equation}
If only part of the canonical edges are verified, each verified edge contributes weak support to its two endpoint vertex pairs:
\begin{equation}
\label{Eq_3.7}
E^w_{i_b^q i_b^c}
\leftarrow
E^w_{i_b^q i_b^c}+1,
\quad
b\in\partial e_a,\ 
a\in\mathcal{A}_{\tau_q,\tau_c},
\quad
\mathrm{if}\ 
0<|\mathcal{A}_{\tau_q,\tau_c}|<3.
\end{equation}

After all triangle pairs have been processed, $\mathbf{E}^{s}$ records repeated complete triangle-level support, while $\mathbf{E}^{w}$ records repeated partial edge-level support. A vertex pair is retained if it receives at least one strong support or satisfies the weak-support acceptance rule:
\begin{equation}
\label{Eq_3.8}
\mathcal{P}_{qc}
=
\{
(m,n)
\mid
E^s_{mn}>0
\ \mathrm{or}\
E^w_{mn} > \tau_w
\},
\end{equation}
where $\tau_w$ is the weak-support threshold.

For each retained vertex pair, the topology-derived reliability weight is defined as
\begin{equation}
\label{Eq_3.9}
\omega_{mn}
=
1
+
\alpha_s\log(1+E^s_{mn})
+
\alpha_w\log(1+E^w_{mn}),
\quad
(m,n)\in\mathcal{P}_{qc},
\end{equation}
where $\alpha_s$ and $\alpha_w$ control the contributions of strong and weak supports, respectively, with $\alpha_s>\alpha_w$.

The 3-D position component of trunk landmark $\mathbf{v}_m^\mu$ is denoted by $\mathbf{p}_m^\mu$, where $\mu\in\{q,c\}$. The retained vertex pairs are finally converted into the weighted vertex correspondence set
\begin{equation}
\label{Eq_3.10}
\mathcal{C}_{qc}
=
\{
(\mathbf{p}_m^q,\mathbf{p}_n^c,\omega_{mn})
\mid
(m,n)\in\mathcal{P}_{qc}
\}
=
\{(\mathbf{p}_k^q,\mathbf{p}_k^c,\omega_k)\}_{k=1}^{K}.
\end{equation}
This correspondence set is used as the input to the decoupled robust pose estimation module.

\begin{algorithm}[t]
\caption{Hierarchical Vertex Correspondence Construction}
\label{alg:alg1}
\small
\begin{algorithmic}[1]
\STATE \textbf{Input:} Query descriptor $\mathcal{D}_q$; candidate descriptor $\mathcal{D}_c$
\STATE \textbf{Output:} Weighted vertex correspondence set $\mathcal{C}_{qc}$

\STATE $\mathcal{C}_{qc}\gets\varnothing$, 
$\mathbf{E}^{s}\gets\mathbf{0}$, 
$\mathbf{E}^{w}\gets\mathbf{0}$

\STATE $s_e\gets S_{\cap}(\mathbf{h}_q^e,\mathbf{h}_c^e)$, 
$s_r\gets S_{\cap}(\mathbf{h}_q^r,\mathbf{h}_c^r)$

\IF{$s_e<\tau_s^e$ \textbf{or} $s_r<\tau_s^r$}
    \STATE \textbf{return} $\mathcal{C}_{qc}$
\ENDIF

\FOR{each triangle pair $(\tau_q,\tau_c)\in\mathcal{T}_q\times\mathcal{T}_c$}
    \STATE $\mathcal{A}_{\tau_q,\tau_c}\gets\varnothing$
    \STATE obtain canonical vertex indices $\mathcal{I}_{\tau_q}$ and $\mathcal{I}_{\tau_c}$

    \FOR{$a=1$ to $3$}
        \STATE let $(u_a,v_a)$ be the endpoint positions of the $a$-th canonical edge
        \STATE $c_a \gets 
        \operatorname{Cons}_{\ell}(\ell_a^q,\ell_a^c)
        \wedge
        \operatorname{Cons}_{r}(r_{u_a}^q,r_{u_a}^c)
        \wedge
        \operatorname{Cons}_{r}(r_{v_a}^q,r_{v_a}^c)$
        \IF{$c_a$}
            \STATE $\mathcal{A}_{\tau_q,\tau_c}\gets\mathcal{A}_{\tau_q,\tau_c}\cup\{a\}$
        \ENDIF
    \ENDFOR

    \IF{$|\mathcal{A}_{\tau_q,\tau_c}|=3$}
        \FOR{$b=1$ to $3$}
            \STATE $E^s_{i_b^q i_b^c}\gets E^s_{i_b^q i_b^c}+1$
        \ENDFOR
    \ELSE
        \FOR{each $a\in\mathcal{A}_{\tau_q,\tau_c}$}
            \STATE let $(u_a,v_a)$ be the endpoint positions of the $a$-th canonical edge
            \STATE $E^w_{i_{u_a}^q i_{u_a}^c}\gets E^w_{i_{u_a}^q i_{u_a}^c}+1$
            \STATE $E^w_{i_{v_a}^q i_{v_a}^c}\gets E^w_{i_{v_a}^q i_{v_a}^c}+1$
        \ENDFOR
    \ENDIF
\ENDFOR

\FOR{each vertex pair $(m,n)\in\mathcal{V}_q\times\mathcal{V}_c$}
    \STATE $\chi_{mn}\gets
    \big(E^s_{mn}>0\big)\vee\big(E^w_{mn}>\tau_w\big)$
    \IF{$\chi_{mn}$}
        \STATE $\omega_{mn}\gets
        1+\alpha_s\log(1+E^s_{mn})
        +\alpha_w\log(1+E^w_{mn})$
        \STATE $\mathcal{C}_{qc}\gets
        \mathcal{C}_{qc}\cup\{(\mathbf{p}_m^q,\mathbf{p}_n^c,\omega_{mn})\}$
    \ENDIF
\ENDFOR

\STATE \textbf{return} $\mathcal{C}_{qc}$
\end{algorithmic}
\end{algorithm}

\subsection{Reliability-Weighted Decoupled Robust Pose Estimation}

Given the weighted vertex correspondence set
$\mathcal{C}_{qc}=\{(\mathbf{p}_k^q,\mathbf{p}_k^c,\omega_k)\}_{k=1}^{K}$,
this module estimates the relative transformation from the query submap to the candidate submap. Robust costs and decoupled pose estimation have been widely used in global registration, as represented by TEASER \cite{yangTEASERFastCertifiable2021} and Quatro \cite{limSingleCorrespondenceEnough2022}. In this study, the topology-derived reliability weights from the front-end are incorporated into a reliability-weighted truncated least-squares formulation:
\begin{equation}
\label{Eq_3.11}
\begin{aligned}
\min_{\theta,\mathbf{t}_{qc}}
&\sum_{k=1}^{K}
\omega_k
\rho_{\beta}
\left(
\left\|
\mathbf{p}_k^c
-
\mathbf{R}_z(\theta)\mathbf{p}_k^q
-
\mathbf{t}_{qc}
\right\|_2
\right),\\
&\rho_{\beta}(r)=\min(r^2,\beta^2),
\end{aligned}
\end{equation}
where $\rho_{\beta}(\cdot)$ suppresses large residuals, and $\omega_k$ transfers the trunk-topology reliability from correspondence construction to pose estimation. Since the submaps are gravity-aligned, roll and pitch have been compensated by the front-end SLAM system. The remaining transformation is therefore estimated in an XOY/H decoupled manner: yaw rotation is first estimated from horizontal translation-invariant measurements, horizontal translation is then estimated in the XOY plane, and elevation translation is finally estimated along the height direction.

\noindent\textbf{Horizontal rotation estimation. }The horizontal projections of the matched trunk positions are denoted by $\mathbf{p}_{k,xy}^q$ and $\mathbf{p}_{k,xy}^c$. Under the gravity-aligned assumption, the rotation between two submaps is restricted to the yaw angle $\theta$, whose horizontal rotation matrix is
\begin{equation}
\label{Eq_3.12}
\mathbf{R}_{xy}(\theta)=
\begin{bmatrix}
\cos\theta & -\sin\theta\\
\sin\theta & \cos\theta
\end{bmatrix}.
\end{equation}

Directly estimating rotation and translation from highly contaminated vertex correspondences is vulnerable to outliers. To remove the translation term, translation-invariant measurements are constructed from correspondence pairs. For a correspondence pair $(a,b)$, the horizontal displacement vectors and the pairwise reliability weight are defined as
\begin{equation}
\label{Eq_3.13}
\mathbf{d}_{ab}^{q}
=
\mathbf{p}_{a,xy}^{q}
-
\mathbf{p}_{b,xy}^{q},
\quad
\mathbf{d}_{ab}^{c}
=
\mathbf{p}_{a,xy}^{c}
-
\mathbf{p}_{b,xy}^{c},
\quad
\omega_{ab}
=
\sqrt{\omega_a\omega_b}.
\end{equation}
The translation component is canceled in these displacement vectors, so the yaw angle can be estimated by aligning the horizontal translation-invariant measurements. The reliability-weighted robust rotation objective is formulated as
\begin{equation}
\label{Eq_3.14}
\theta^\star
=
\arg\min_{\theta}
\sum_{(a,b)\in\mathcal{E}}
\omega_{ab}
\rho_{\beta_R}
\left(
\left\|
\mathbf{d}_{ab}^{c}
-
\mathbf{R}_{xy}(\theta)\mathbf{d}_{ab}^{q}
\right\|_2
\right),
\end{equation}
where $\mathcal{E}$ is the set of translation-invariant measurement pairs. The pairwise weight $\omega_{ab}$ assigns higher confidence to measurements constructed from two reliable vertex correspondences, while the truncated least-squares cost reduces the influence of inconsistent pairs. The resulting one-dimensional non-convex robust problem is solved using GNC \cite{yangGraduatedNonConvexityRobust2020a}. After yaw estimation, correspondences consistent with the estimated rotation are collected into the inlier set $\mathcal{I}_R$ and passed to translation estimation.

\noindent\textbf{Reliability-weighted translation estimation. }After obtaining $\theta^\star$, the rotation-compensated horizontal translation observation of each rotation-consistent correspondence is computed as
\begin{equation}
\label{Eq_3.15}
\boldsymbol{\delta}_k^{xy}
=
\mathbf{p}_{k,xy}^{c}
-
\mathbf{R}_{xy}(\theta^\star)\mathbf{p}_{k,xy}^{q},
\quad
k\in\mathcal{I}_R .
\end{equation}
For correct correspondences, these observations should concentrate around the true horizontal translation. However, residual false matches may still remain after rotation estimation. Therefore, the horizontal translation is estimated by another reliability-weighted truncated least-squares problem:
\begin{equation}
\label{Eq_3.16}
\mathbf{t}_{xy}^\star
=
\arg\min_{\mathbf{t}_{xy}}
\sum_{k\in\mathcal{I}_R}
\omega_k
\rho_{\beta_{xy}}
\left(
\left\|
\boldsymbol{\delta}_k^{xy}
-
\mathbf{t}_{xy}
\right\|_2
\right).
\end{equation}
This formulation treats the horizontal translation as a robust weighted location estimation problem, where reliable vertex correspondences have larger influence and large residuals are truncated.

After horizontal translation estimation, the retained translation-consistent inlier set is denoted as $\mathcal{I}_T$. The elevation component is then estimated independently from the height differences of these inliers. For each $k\in\mathcal{I}_T$, the elevation translation observation is $\delta_k^z=p_{k,z}^{c}-p_{k,z}^{q}$, and the elevation translation is estimated as
\begin{equation}
\label{Eq_3.17}
t_z^\star
=
\arg\min_{t_z}
\sum_{k\in\mathcal{I}_T}
\omega_k
\rho_{\beta_z}
\left(
\left|
\delta_k^z
-
t_z
\right|
\right).
\end{equation}
The final relative transformation is given by
$\mathbf{R}_{qc}^\star=\mathbf{R}_z(\theta^\star)$ and
$\mathbf{t}_{qc}^\star=((\mathbf{t}_{xy}^\star)^\top,t_z^\star)^\top$, which is used as the loop closure constraint between the query and candidate submaps.

\section{Experiments} \label{section:4} 
\subsection{Experimental Setup}
\subsubsection{Datasets and Ground-Truth Construction}

The proposed method was evaluated on both simulated and real-world datasets. The simulated data were used to assess the loop closure detection and relative pose estimation performance of different methods under controlled conditions, while the real-world data were used to further examine their applicability and stability in actual forest understory environments. All experiments took gravity-aligned local point cloud submaps as input, and the relative pose estimated between each submap pair was compared with its ground-truth counterpart.

\begin{figure*}[!t]
\centering
\includegraphics[width=0.92\textwidth]{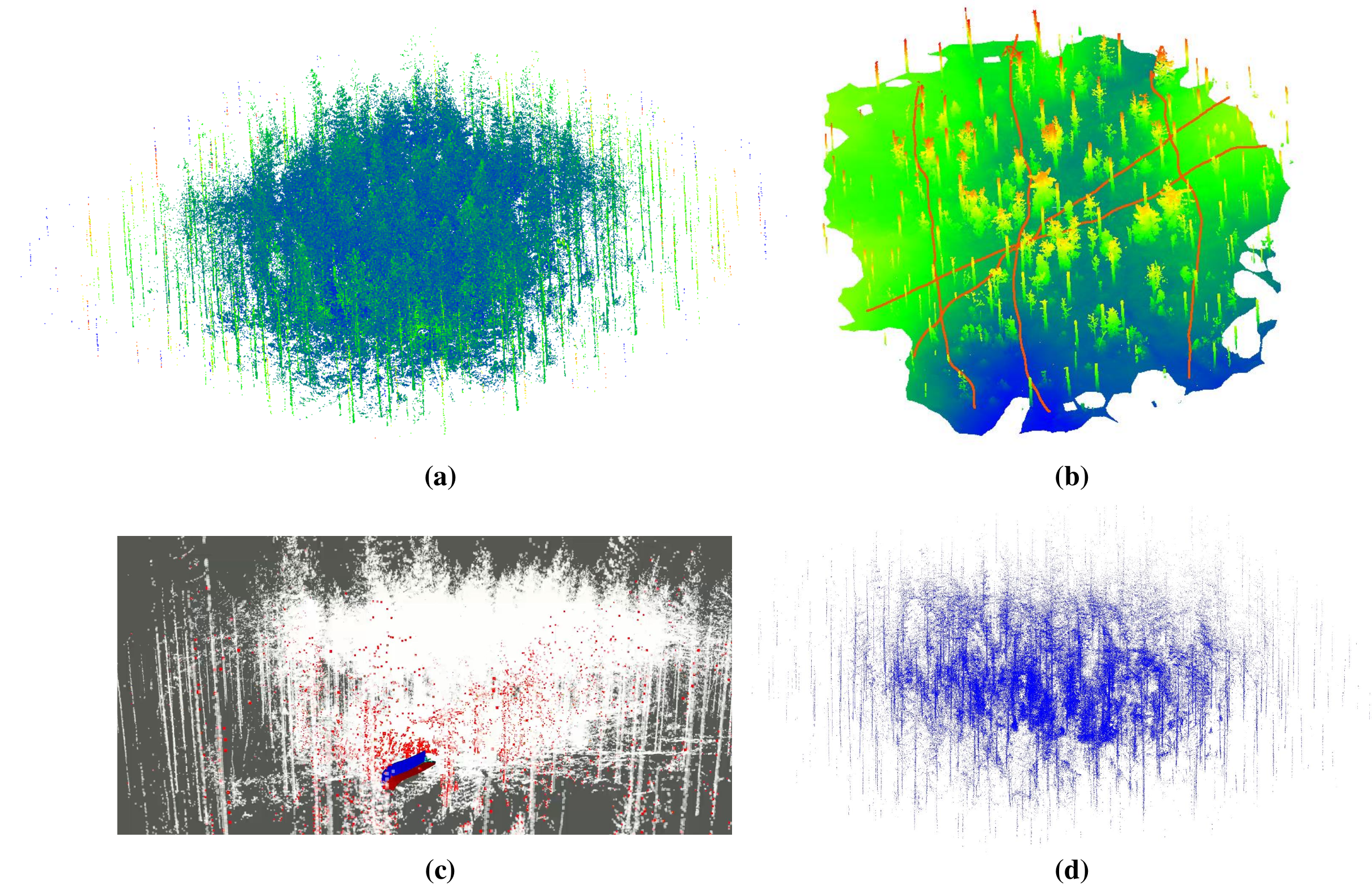}
\caption{Simulated forest dataset generation process. 
        (a) Original WHU-TLS point cloud. 
        (b) Multiple trajectories generated by PCT-Planner. 
        (c) LiDAR simulation using MARSIM. 
        (d) Map reconstructed by FAST-LIO2.}
\label{fig_2}
\end{figure*}

The simulation experiments were conducted using a forest plot. Gravity-aligned local point cloud submaps and their reference poses were generated through simulated path planning, LiDAR scanning simulation, and mapping processing. The complete data generation procedure is further described in Section~\ref{section:simulated} with reference to Fig.~\ref{fig_2}.

\begin{figure*}[!t]
\centering
\includegraphics[width=0.92\textwidth]{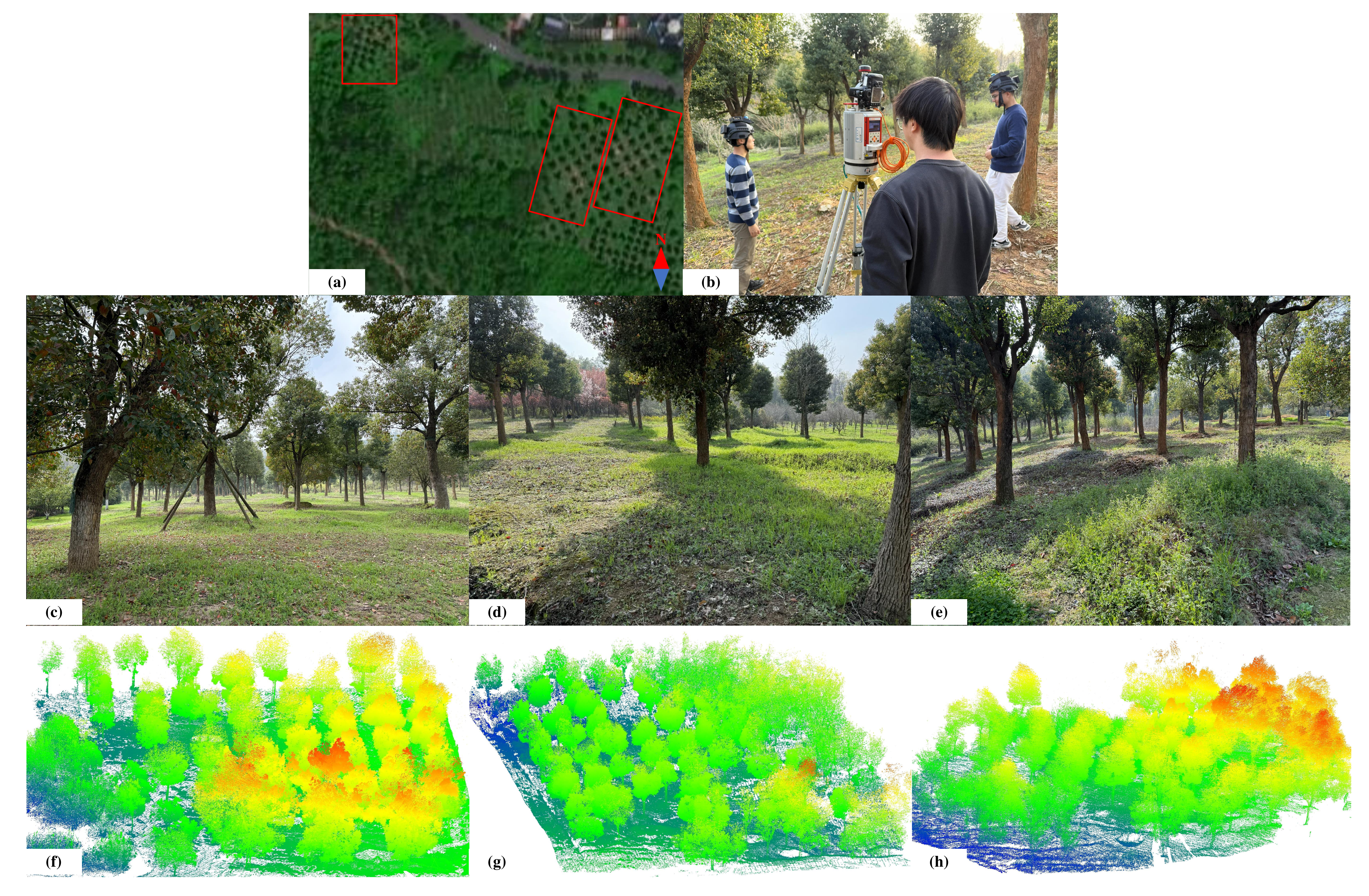}
\caption{Overview of the real-world forest plots and TLS data acquisition. 
        (a) Locations of the three experimental plots. 
        (b) Field data acquisition. 
        (c)--(e) Field views of the three plots. 
        (f)--(h) Corresponding TLS point clouds.}
\label{fig_3}
\end{figure*}

The real-world experiments were conducted in three forest plots. Further details of the data acquisition setup and forest plots are provided in Section~\ref{section:realworld} with reference to Fig.~\ref{fig_3}.

Let the ground-truth pose of the $i$-th submap in the reference coordinate system be
\begin{equation}
\label{Eq_4.1}
\mathbf{T}_i^\star
=
\begin{bmatrix}
\mathbf{R}_i^\star & \mathbf{t}_i^\star\\
\mathbf{0}^\top & 1
\end{bmatrix}
\end{equation}

where $\mathbf{R}_i^\star\in SO(3)$ and $\mathbf{t}_i^\star\in\mathbb{R}^3$. For a query submap $i$ and a historical submap $j$, if the relative transformation is defined as mapping coordinates from the frame of submap $i$ to that of submap $j$, the ground-truth relative pose is given by
\begin{equation}
\label{Eq_4.2}
\mathbf{T}_{ij}^\star
=
(\mathbf{T}_j^\star)^{-1}
\mathbf{T}_i^\star
=
\begin{bmatrix}
\mathbf{R}_{ij}^\star & \mathbf{t}_{ij}^\star\\
\mathbf{0}^\top & 1
\end{bmatrix}
\end{equation}

The relative pose estimated by each method is denoted as
\begin{equation}
\label{Eq_4.3}
\hat{\mathbf{T}}_{ij}
=
\begin{bmatrix}
\hat{\mathbf{R}}_{ij} & \hat{\mathbf{t}}_{ij}\\
\mathbf{0}^\top & 1
\end{bmatrix}
\end{equation}

All accuracy metrics reported below are computed from the discrepancy between $\hat{\mathbf{T}}_{ij}$ and $\mathbf{T}_{ij}^\star$.

\subsubsection{Comparison Methods}
STD~\cite{yuanSTDStableTriangle2023}, Quatro++~\cite{limQuatroRobustGlobal2024}, and GROR~\cite{yanNewOutlierRemoval2022} were selected as baseline methods. STD is a triangle-descriptor-based loop closure detection method that constructs STDesc descriptors from planar and corner features for loop candidate retrieval and relative pose estimation. Quatro++ is a robust global registration method based on local feature correspondences. It establishes tentative correspondences through point cloud downsampling, normal estimation, and FPFH descriptor matching, followed by robust pose estimation. GROR also relies on local feature correspondences but performs robust initial registration using geometric consistency constraints, generally providing strong outlier rejection capability.

Unlike STD, the proposed method does not rely on local geometric features such as planes and corners, but directly exploits the more stable spatial topology of tree trunks in forest environments. In contrast to Quatro++ and GROR, the proposed method constructs sparse weighted vertex correspondences at the front end and propagates their reliability to the back-end pose estimator as prior weights.

To further investigate the influence of different back-end solvers on registration performance, three robust pose estimation methods, namely FGR~\cite{zhouFastGlobalRegistration2016}, TEASER~\cite{yangTEASERFastCertifiable2021}, and Quatro~\cite{limSingleCorrespondenceEnough2022}, were compared. In addition, a unit-weight ablation was conducted to evaluate the contribution of correspondence reliability weights to back-end pose estimation. Specifically, while keeping the front-end correspondence construction process unchanged, all tentative correspondences were assigned identical weights, and the resulting performance was compared with that of the complete method.

\subsubsection{Evaluation Metrics}

We evaluated each method in terms of pose accuracy and computational time. For each accepted submap pair, the estimated relative pose $(\hat{\mathbf{R}}_{ij},\hat{\mathbf{t}}_{ij})$ was compared with the ground-truth relative pose $(\mathbf{R}_{ij}^{\star},\mathbf{t}_{ij}^{\star})$.

The total rotation error was computed as the geodesic distance on $SO(3)$:
\begin{equation}
\label{Eq_4.4}
e_R
=
\cos^{-1}
\left(
\frac{
\mathrm{trace}\left((\mathbf{R}_{ij}^{\star})^\top \hat{\mathbf{R}}_{ij}\right)-1
}{2}
\right)
\cdot
\frac{180}{\pi}.
\end{equation}
The argument of the inverse cosine was clipped to $[-1,1]$ to avoid numerical errors.

The roll, pitch, and yaw errors were obtained from the ZYX Euler decomposition of the rotation residual $(\mathbf{R}_{ij}^{\star})^\top \hat{\mathbf{R}}_{ij}$. Translation errors were computed from the residual $\hat{\mathbf{t}}_{ij}-\mathbf{t}_{ij}^{\star}$, including the component-wise errors along the $x$-, $y$-, and $z$-axes and the total Euclidean translation error.

For each dataset, we reported the mean errors over all valid test samples. The RMSEs of the total rotation and translation errors were also reported to reflect the effect of large residuals.

Computational time was reported as the average time per submap. In the main comparison experiments, we reported the descriptor or feature construction time $T_{\mathrm{desc}}$, the loop closure retrieval and optimization time $T_{\mathrm{loop}}$, and the total time $T_{\mathrm{total}}=T_{\mathrm{desc}}+T_{\mathrm{loop}}$. In the ablation and back-end comparison experiments, the solver time $T_{\mathrm{solver}}$ was additionally reported.

\subsection{Experiments on Simulated Data} \label{section:simulated}
The experiments on simulated data were conducted to evaluate the performance of different methods in loop closure detection and global registration of forest understory point cloud submaps under conditions with known ground-truth poses. Compared with real-world data, simulated data provide more complete reference pose relationships, facilitating quantitative evaluation of the relative poses estimated by different methods. Moreover, because the simulation reproduces the scanning characteristics of a low-cost LiDAR and the trajectories of a mobile platform, the resulting data can reflect, to a certain extent, the effects of sparse forest point clouds, repetitive trunk structures, and local occlusions on loop closure detection and registration accuracy.

As shown in Fig.~\ref{fig_2}, the simulated data were generated from a forest plot in the WHU-TLS dataset\cite{dongRegistrationLargescaleTerrestrial2020a}. The core region of the plot covers approximately $50~\mathrm{m} \times 65~\mathrm{m}$ and contains a representative spatial distribution of tree trunks. First, multiple simulated acquisition trajectories were generated within the plot using PCT-Planner~\cite{yangEfficientGlobalNavigational2024} to reproduce different observation paths of a mobile platform in a forest understory environment. Subsequently, MARSIM~\cite{kongMARSIMLightWeightPointRealistic2023} was used to simulate data acquisition by a mobile platform equipped with a MID-360 LiDAR. This simulation retained the spatial structure of the original TLS forest plot while introducing the scanning characteristics of a low-cost LiDAR. Finally, the simulated LiDAR measurements were processed using FAST-LIO2~\cite{xuFASTLIO2FastDirect2022} to obtain gravity-aligned point cloud frames and their corresponding poses. Consecutive point cloud frames were accumulated into local submaps using a fixed temporal window, and the ground-truth relative transformations between submaps were computed from their poses in the reference coordinate system.

\begin{figure*}[!t]
\centering
\includegraphics[width=0.92\textwidth,
                height=0.82\textheight,
                keepaspectratio]{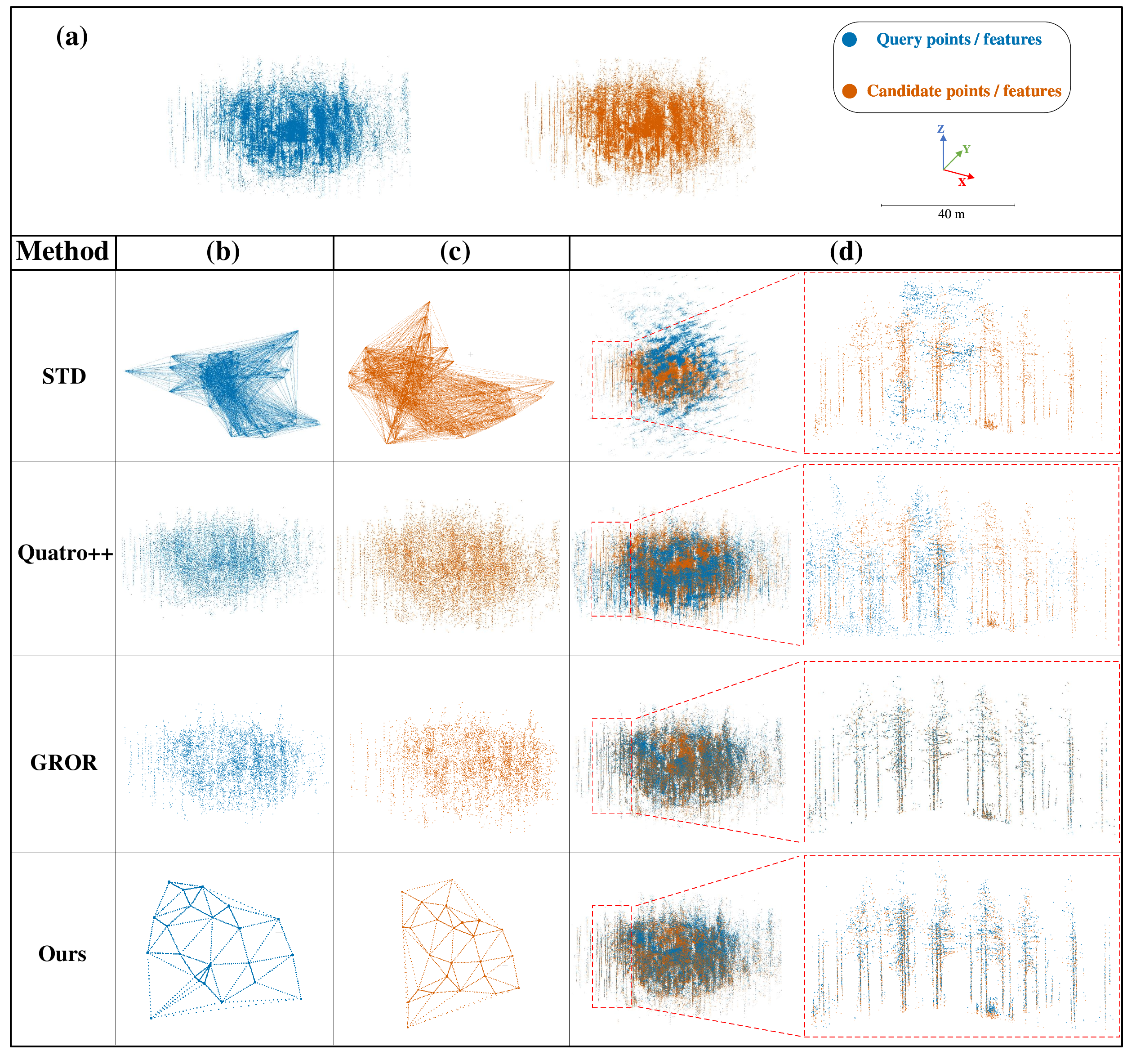}
\caption{Qualitative comparison of feature representations and registration results on the simulated data.
        (a) Input query and candidate submaps.
        (b) and (c) Feature representations constructed from the query and candidate submaps, respectively, by STD, Quatro++, GROR, and the proposed method.
        (d) Registration results of the corresponding methods, where the red dashed boxes indicate the selected local regions and the enlarged views on the right highlight the alignment of individual trunks.
        The color legend in (a) applies to all subfigures: blue and orange denote the query and candidate points or features, respectively.}
\label{fig_4}
\end{figure*}

Fig.~\ref{fig_4} presents a representative qualitative comparison on the simulated forest dataset. As shown in Fig.~\ref{fig_4}(a), the query and candidate submaps contain numerous vertically distributed trunks with similar local arrangements and spacing patterns. Such repetitive structures reduce the distinctiveness of local geometric cues and increase the probability of introducing ambiguous correspondences. The feature representations generated by different methods are shown in Fig.~\ref{fig_4}(b) and Fig.~\ref{fig_4}(c). STD constructs dense triangle-based descriptors from local planar and corner features; however, the query and candidate descriptor structures exhibit substantial geometric inconsistency, indicating that these local primitives are not sufficiently repeatable in the simulated forest scene. In contrast, the proposed method represents each submap using a sparse Delaunay topology built from stable trunk landmarks. The resulting query and candidate topologies preserve similar large-scale structural patterns while substantially reducing descriptor redundancy.

The corresponding registration results are shown in Fig.~\ref{fig_4}(d), where the enlarged regions highlight the alignment of individual trunks. STD produces evident rotational and translational misalignment, which is consistent with the poor correspondence between its triangle-based feature structures. Quatro++ improves the global orientation but still exhibits noticeable residual offsets between the query and candidate trunks. GROR achieves substantially better registration results, indicating that its geometric consistency verification can effectively suppress incorrect correspondences and recover an accurate relative transformation. The proposed method also achieves close overlap between corresponding trunk structures in both the global and enlarged views, with registration accuracy comparable to that of GROR. These results demonstrate that trunk-topology-guided vertex correspondences can provide reliable geometric constraints for loop closure registration in repetitive forest environments.

\begin{table*}[!t]
\caption{Statistical Results of Simulation Data}
\label{tab:table1}
\centering
\scriptsize
\setlength{\tabcolsep}{2.5pt}

\begin{tabular}{|c|cccc|cccc|cc|ccc|}
\hline

\multirow{2}{*}{Method}
& \multicolumn{4}{c|}{Average Rotation Error (deg)}
& \multicolumn{4}{c|}{Average Translation Error (m)}
& \multicolumn{2}{c|}{RMSE}
& \multicolumn{3}{c|}{Average Processing Time (ms)}
\\ \cline{2-14}

& Roll & Pitch & Yaw & Total
& X-axis & Y-axis & Z-axis & Total
& RMSE-R & RMSE-T
& Desc./Feat. & Loop \& Opt. & Total
\\ \hline

STD
& 35.136 & 8.504 & 72.785 & 80.531
& 11.144 & 19.559 & 2.551 & 24.566
& 98.870 & 27.111
& 15.407 & 142.527 & 157.934
\\ \hline

Quatro++
& 0.527 & 0.786 & 84.552 & 84.552
& 10.979 & 15.090 & 2.338 & 19.069
& 93.086 & 20.333
& 25.030 & 26921.501 & 26946.531
\\ \hline

GROR
& 0.281 & 0.442 & 0.106 & 0.583
& 0.049 & 0.040 & 0.141 & 0.169
& 0.629 & 0.192
& 616.707 & 23727.401 & 24344.108
\\ \hline

Ours
& 0.596 & 0.500 & 0.098 & 0.889
& 0.027 & 0.044 & 0.436 & 0.445
& 1.066 & 0.537
& 200.427 & 13.862 & 214.289
\\ \hline

\end{tabular}

\vspace{2pt}
\begin{minipage}{\textwidth}
\footnotesize
\textit{Note:} RMSE-R and RMSE-T denote the root mean square
errors of the total rotation and translation errors, respectively.
Desc./Feat., Loop \& Opt., and Total denote the descriptor or feature
construction time, loop closure retrieval and optimization time,
and total processing time, respectively.
\end{minipage}
\end{table*}

Table~\ref{tab:table1} reports the pose estimation accuracy and average processing time of STD, Quatro++, GROR, and the proposed method on the simulated dataset. Overall, STD and Quatro++ produce large horizontal rotation and translation errors in the simulated forest environment, whereas GROR and the proposed method achieve substantially higher registration accuracy. However, GROR obtains its high accuracy at considerable computational cost, while the proposed method significantly reduces the average processing time without substantially compromising accuracy.

STD produces clear registration failures in the simulated environment, with mean total rotation and translation errors of $80.531^{\circ}$ and $24.566~\mathrm{m}$, respectively. The yaw error constitutes the dominant component of its rotation error, indicating that STD has difficulty reliably determining horizontal rotation in forest environments with repetitive trunk structures. The RMSE-R and RMSE-T values further show that STD exhibits not only large mean errors but also severe registration failures for some submap pairs. This suggests that triangle descriptors constructed from planar and corner features are susceptible to point cloud sparsity, local occlusion, and foliage clutter, resulting in incorrect candidate matches among repetitive trunk structures. Quatro++ performs well in the roll and pitch directions, but its yaw and total translation errors remain large. Its mean yaw error reaches $84.552^{\circ}$, while its mean total translation error is $19.069~\mathrm{m}$. These results indicate that, under gravity-aligned conditions, a general local-feature-based registration method can effectively constrain the pose degrees of freedom associated with the gravity direction, but still has difficulty resolving horizontal rotation ambiguities in forest environments. Because many tree trunks exhibit similar local geometric structures, tentative correspondences established using local descriptors such as FPFH may contain structurally incorrect matches. Consequently, large yaw and horizontal translation errors may remain even when a robust registration strategy is employed at the back end. GROR achieves the highest mean registration accuracy on the simulated dataset, with mean total rotation and translation errors of $0.583^{\circ}$ and $0.169~\mathrm{m}$, respectively. Its RMSE values also indicate relatively stable registration performance. These results demonstrate that robust global registration based on geometric consistency constraints can effectively suppress outliers and recover accurate poses when the tentative correspondences are sufficiently reliable. Nevertheless, GROR incurs a substantial computational cost, requiring approximately $24.3~\mathrm{s}$ per submap on average. Such a processing time makes it difficult to satisfy the real-time requirements of online edge platforms or collaborative data acquisition using multiple devices. The proposed method achieves a favorable balance between accuracy and efficiency on the simulated dataset. Its mean total rotation and translation errors are $0.889^{\circ}$ and $0.445~\mathrm{m}$, respectively. Although its overall accuracy is slightly lower than that of GROR, it substantially outperforms STD and Quatro++. In particular, the yaw error of the proposed method is $0.098^{\circ}$, which is comparable to the $0.106^{\circ}$ obtained by GROR. This result indicates that horizontal rotation estimation based on TIMs can effectively recover the dominant rotational degree of freedom between forest submaps. The horizontal translation errors also remain low, demonstrating that the horizontal spatial topology formed by trunk centers provides stable constraints for inter-submap translation estimation.

It should be noted that the translation error of the proposed method along the $z$-axis is noticeably larger than its horizontal translation errors. This behavior is mainly related to the simulation data generation process. To ensure real-time point cloud rendering, the original TLS forest point cloud was downsampled before simulation. In addition, the MID-360 LiDAR employs a non-repetitive scanning pattern, resulting in relatively sparse ground and trunk-base points during simulated mobile data acquisition. Consequently, trunk-base elevation estimation is more susceptible to point cloud sparsity and local missing data, leading to a comparatively larger vertical translation error. These results indicate that the principal advantages of the proposed method on the simulated dataset lie in horizontal rotation and translation estimation, whereas vertical translation estimation is more sensitive to the quality of ground and trunk-base points.

Regarding computational efficiency, STD requires the least time for descriptor construction but provides poor registration accuracy. The computational costs of Quatro++ and GROR are dominated by loop closure retrieval and robust registration. In comparison, the representation construction time of the proposed method is higher than that of STD because it involves trunk landmark extraction, circular cross-section fitting, Delaunay topology construction, and vertex support accumulation. Nevertheless, its loop closure retrieval and optimization time is only $13.862~\mathrm{ms}$, which is substantially lower than those of the baseline methods. The proposed method requires an average total processing time of $214.289~\mathrm{ms}$, which is considerably shorter than those of Quatro++ and GROR. Together with the accuracy and RMSE results, these observations demonstrate that compressing the original point clouds into sparse trunk landmarks and compact topological relationships substantially reduces the computational cost of correspondence construction and pose estimation while maintaining high accuracy and stability.

In summary, the simulated-data experiments demonstrate that STD and Quatro++ are susceptible to repetitive trunk structures and local feature ambiguities, resulting in large yaw and horizontal translation errors. GROR achieves high registration accuracy but incurs excessive computational cost. By integrating trunk-topology representation, vertex correspondence construction, and reliability-weighted XOY/H-decoupled robust estimation, the proposed method achieves horizontal rotation and translation accuracy comparable to that of GROR while substantially reducing the processing time. The $z$-axis error observed in the simulated data further indicates that vertical translation estimation is sensitive to the quality of ground and trunk-base points, which is further examined in the real-world experiments.

\subsection{Experiments on Real-World Data} \label{section:realworld}
The real-world experiments were conducted to further evaluate the applicability of the proposed method in actual forest understory environments. Compared with simulated data, real-world forest point clouds are affected by additional factors, including terrain variations, trunk occlusions, foliage clutter, local point cloud incompleteness, motion disturbances of the mobile platform, and viewpoint differences among acquisition trajectories. Therefore, these experiments provide a more direct evaluation of loop closure detection and global registration performance under practical forest inventory conditions.

As shown in Fig.~\ref{fig_3}, the real-world data were collected from three forest plots in Ma'anshan Forest Park, Wuhan, China, denoted as Real Plot 1, Real Plot 2, and Real Plot 3. These plots differ in stand density, trunk distribution, terrain variation, and understory clutter, enabling evaluation under diverse forest conditions. In each plot, multiple WHU-Helmet\cite{liWHUHelmetHelmetbasedMultisensor2023} mobile laser scanning systems equipped with MID-360 LiDAR sensors were used to collect understory point clouds along different trajectories. Meanwhile, a RIEGL VZ-400 terrestrial laser scanner was used to acquire high-accuracy TLS point clouds as the ground-truth reference in a common coordinate system. Specifically, the TLS point clouds were incorporated into FAST-LIO2\cite{xuFASTLIO2FastDirect2022} as reference maps to constrain the point cloud frames generated by the mobile platforms, thereby obtaining the ground-truth pose of each local submap in the common reference coordinate system.

\begin{figure*}[!t]
\centering
\includegraphics[width=0.92\textwidth,
                height=0.82\textheight,
                keepaspectratio]{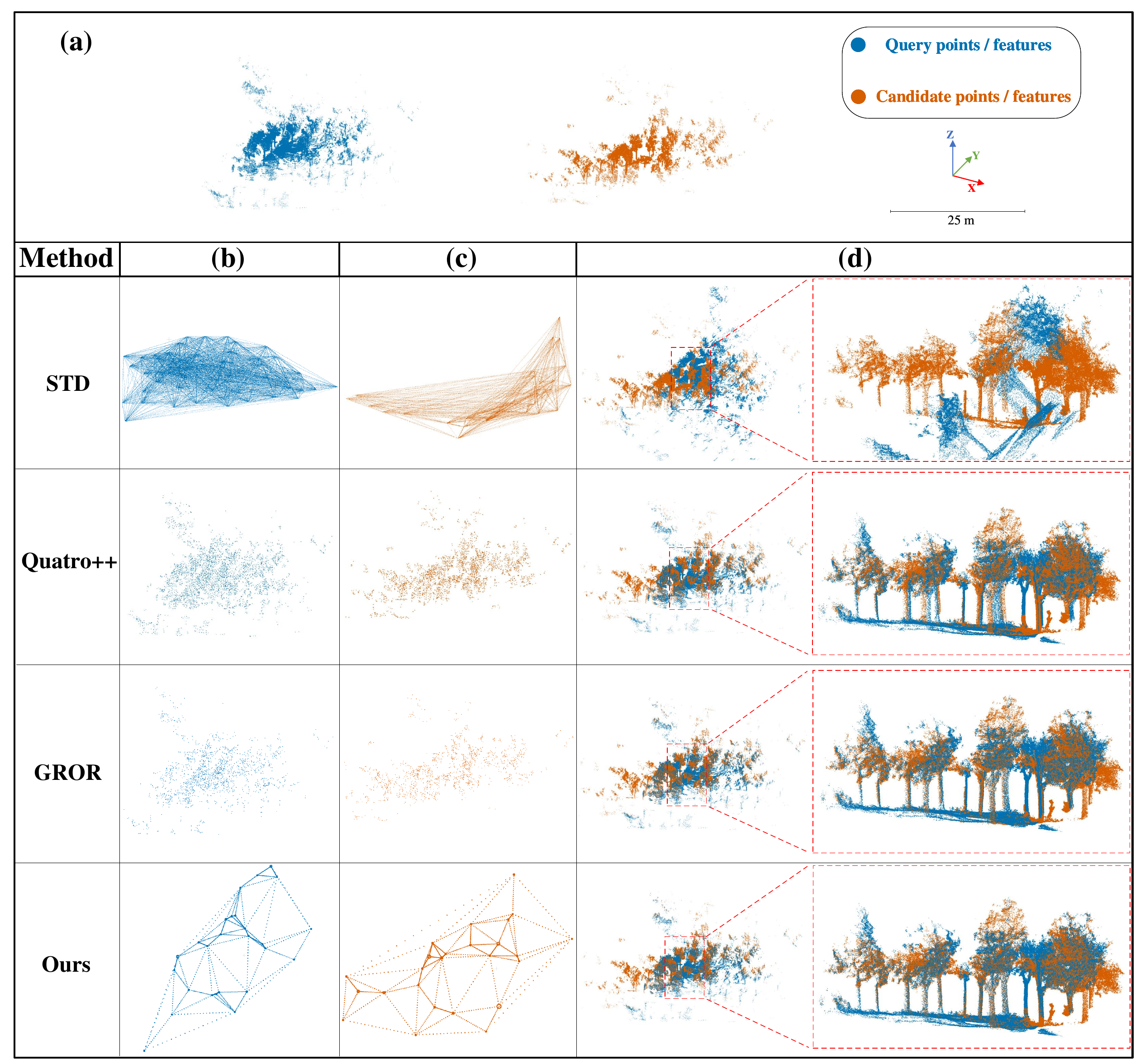}
\caption{Qualitative comparison of feature representations and registration results on the real-world data.
        (a) Input query and candidate submaps.
        (b) and (c) Feature representations constructed from the query and candidate submaps, respectively, by STD, Quatro++, GROR, and the proposed method.
        (d) Registration results of the corresponding methods, where the red dashed boxes indicate the selected local regions and the enlarged views on the right highlight the alignment of tree structures.
        The color legend in (a) applies to all subfigures: blue and orange denote the query and candidate points or features, respectively.}
\label{fig_5}
\end{figure*}

Fig.~\ref{fig_5} presents a representative qualitative comparison on the real-world forest data. As shown in Fig.~\ref{fig_5}(a), the query and candidate submaps exhibit only partial overlap and contain substantial occlusion, nonuniform point density, and locally missing structures. These factors introduce additional uncertainty into feature extraction and correspondence construction, making the real-world scene a more demanding test of registration robustness.

The feature representations in Fig.~\ref{fig_5}(b) and Fig.~\ref{fig_5}(c) reveal clear differences among the evaluated methods. STD constructs dense triangle-based descriptor structures from local geometric features, but the query and candidate triangle networks differ considerably and contain numerous long-range connections across sparse or incomplete regions. Quatro++ and GROR retain sparser point-level representations, whose spatial distributions are still influenced by occlusion and uneven sampling. In contrast, the proposed method consistently extracts trunk-topology vertices from both submaps and constructs compact Delaunay structures that preserve similar spatial organization despite incomplete observations.

The corresponding registration results are shown in Fig.~\ref{fig_5}(d), where the enlarged views highlight the alignment of local tree structures. STD produces substantial rotational and translational misalignment, while Quatro++ recovers the overall orientation but retains evident local offsets. GROR achieves accurate alignment in the principal overlapping region, demonstrating the effectiveness of its robust geometric consistency verification on real-world forest point clouds. The proposed method also produces close overlap between the query and candidate submaps in both the global and enlarged views, with registration accuracy comparable to that of GROR. These results indicate that the proposed trunk-topology representation remains sufficiently stable under partial overlap, occlusion, nonuniform sampling, and local structural incompleteness, thereby providing reliable vertex correspondences for real-world forest loop closure registration.

\begin{table*}[!t]
\caption{Statistical Results of Real Data}
\label{tab:table2}
\centering
\scriptsize
\setlength{\tabcolsep}{2.5pt}
\begin{tabular}{|c|c|cccc|cccc|cc|ccc|}
\hline
\multirow{2}{*}{Plot}
& \multirow{2}{*}{Method}
& \multicolumn{4}{c|}{Average Rotation Error (deg)}
& \multicolumn{4}{c|}{Average Translation Error (m)}
& \multicolumn{2}{c|}{RMSE}
& \multicolumn{3}{c|}{Average Processing Time (ms)}
\\ \cline{3-15}

&
& Roll & Pitch & Yaw & Total
& X-axis & Y-axis & Z-axis & Total
& RMSE-R (deg) & RMSE-T (m)
& Desc./Feat. & Loop \& Opt. & Total
\\ \hline

\multirow{4}{*}{Real Plot 1}
& STD
& 66.277 & 3.876 & 75.811 & 113.723
& 20.434 & 22.479 & 1.712 & 33.806
& 133.502 & 41.934
& 19.651 & 159.393 & 179.044
\\ \cline{2-15}

& Quatro++
& 0.024 & 0.082 & 27.591 & 27.598
& 8.645 & 8.220 & 0.282 & 12.955
& 54.634 & 23.589
& 18.247 & 3064.512 & 3082.759
\\ \cline{2-15}

& GROR
& 0.181 & 0.347 & 0.095 & 0.422
& 0.037 & 0.031 & 0.109 & 0.130
& 0.509 & 0.163
& 1831.406 & 73558.984 & 75390.390
\\ \cline{2-15}

& Ours
& 0.033 & 0.090 & 0.106 & 0.176
& 0.042 & 0.053 & 0.056 & 0.103
& 0.307 & 0.203
& 375.334 & 14.748 & 390.081
\\ \hline

\multirow{4}{*}{Real Plot 2}
& STD
& 71.623 & 3.640 & 88.600 & 121.780
& 21.372 & 23.138 & 2.138 & 35.057
& 137.804 & 40.267
& 17.572 & 133.294 & 150.867
\\ \cline{2-15}

& Quatro++
& 0.039 & 0.015 & 16.603 & 16.605
& 5.844 & 5.126 & 0.236 & 8.545
& 42.668 & 19.813
& 16.711 & 3037.824 & 3054.535
\\ \cline{2-15}

& GROR
& 0.208 & 0.293 & 0.056 & 0.397
& 0.024 & 0.023 & 0.075 & 0.091
& 0.495 & 0.118
& 1664.655 & 65386.701 & 67051.355
\\ \cline{2-15}

& Ours
& 0.013 & 0.004 & 0.282 & 0.271
& 0.090 & 0.039 & 0.017 & 0.108
& 1.634 & 0.559
& 370.906 & 14.054 & 384.960
\\ \hline

\multirow{4}{*}{Real Plot 3}
& STD
& 45.799 & 3.476 & 68.876 & 86.950
& 17.102 & 20.700 & 2.447 & 29.694
& 112.464 & 36.817
& 22.144 & 308.931 & 331.075
\\ \cline{2-15}

& Quatro++
& 0.089 & 0.046 & 10.542 & 10.552
& 2.869 & 1.885 & 0.128 & 3.745
& 34.200 & 10.953
& 19.987 & 3623.602 & 3643.589
\\ \cline{2-15}

& GROR
& 0.311 & 0.410 & 0.086 & 0.566
& 0.031 & 0.038 & 0.113 & 0.134
& 0.730 & 0.179
& 1832.107 & 69803.826 & 71635.933
\\ \cline{2-15}

& Ours
& 0.060 & 0.010 & 0.110 & 0.125
& 0.034 & 0.032 & 0.035 & 0.067
& 0.275 & 0.117
& 390.713 & 17.128 & 407.841
\\ \hline

\end{tabular}

\vspace{2pt}
\begin{minipage}{\textwidth}
\footnotesize
\textit{Note:} RMSE-R and RMSE-T denote the root mean square
errors of the total rotation and translation errors, respectively.
Desc./Feat., Loop \& Opt., and Total denote the descriptor or feature
construction time, loop closure retrieval and optimization time,
and total processing time, respectively.
\end{minipage}
\end{table*}

Table~\ref{tab:table2} reports the pose estimation accuracy and average processing time of STD, Quatro++, GROR, and the proposed method on the three real-world forest plots. Overall, the error patterns observed on the real-world data are consistent with those obtained on the simulated data. STD and Quatro++ remain susceptible to large yaw and horizontal translation errors, whereas GROR and the proposed method maintain high registration accuracy. The real-world experiments further reveal differences in the stability of these methods across diverse forest conditions.

STD produces large total rotation and translation errors in all three real-world plots. Its total rotation errors range from several tens to more than one hundred degrees, while its total translation errors are on the order of tens of meters. These results indicate that triangle descriptors constructed from planar and corner features remain vulnerable to unstable local geometric structures in real-world forest data. Compared with the simulated environment, foliage occlusion, local point-cloud incompleteness, and terrain variations further reduce the repeatability of planar and corner features. Consequently, incorrect candidate matches remain a significant problem for STD on the real-world data. Quatro++ is generally stable in the roll and pitch directions, but still exhibits large yaw and horizontal translation errors. This observation is consistent with the simulated-data results. In real-world forest plots, many trunks exhibit similar local geometric structures, while viewpoint and occlusion differences exist among acquisition trajectories. Therefore, even with a robust back-end registration strategy, Quatro++ cannot consistently recover accurate yaw and horizontal translation estimates. GROR maintains high registration accuracy across all three real-world plots. Its mean total rotation error, mean total translation error, and RMSE values remain low, demonstrating that geometric consistency constraints provide strong outlier rejection capability in real-world forest data. However, this accuracy is accompanied by considerable computational cost. The average total processing time of GROR is approximately $67$--$75~\mathrm{s}$ per submap across the three plots, with most of the computation spent on local feature matching and large-scale geometric consistency verification. GROR can therefore serve as a high-accuracy reference method, but it is difficult to deploy for real-time processing on online edge platforms or in collaborative data acquisition using multiple devices. The proposed method achieves high pose estimation accuracy in all three real-world plots and provides a more favorable balance between accuracy and computational efficiency. Compared with STD and Quatro++, the proposed method substantially reduces the yaw and horizontal translation errors. Compared with GROR, it achieves comparable or lower mean total rotation and translation errors. In particular, for Real Plot 1 and Real Plot 3, both the mean total rotation and translation errors of the proposed method are lower than those of GROR. These results demonstrate that trunk landmarks, Delaunay topological relationships, and reliability-weighted XOY/H-decoupled robust estimation provide stable constraints in real-world forest environments. 

It should be noted that, although the mean errors of the proposed method remain low in Real Plot 2, its RMSE-R and RMSE-T values are higher than those of GROR because of a small number of erroneous matches in this plot. This observation indicates that weighted vertex correspondences may still be affected by local structural ambiguities in regions with highly repetitive trunk distributions or incomplete observations. Nevertheless, considering both mean errors and computational time, the proposed method maintains strong overall registration performance and substantially outperforms STD and Quatro++. Another noteworthy observation is that the $z$-axis translation errors of the proposed method on the real-world data are generally lower than those on the simulated data. The mean $z$-axis errors in all three real-world plots remain at the centimeter level, whereas the corresponding error on the simulated dataset is noticeably larger. This difference is primarily attributable to the data generation and ground-truth construction procedures. In the real-world plots, direct sensor measurements provide more complete observations of the ground and trunk bases. In contrast, the original TLS forest point cloud was downsampled before simulation, while the non-repetitive scanning pattern of the MID-360 LiDAR further reduced the density of ground and trunk-base points. Consequently, trunk-base elevation estimation on the simulated data is more susceptible to point-cloud sparsity and local incompleteness, resulting in larger vertical translation errors.

Regarding computational efficiency, the proposed method requires approximately $0.38$--$0.41~\mathrm{s}$ per submap on the three real-world plots, which is substantially shorter than the processing times of Quatro++ and GROR. Although its descriptor and feature construction time is higher than that of STD, its loop closure retrieval and optimization time remains on the order of only tens of milliseconds. This result is consistent with the simulated-data experiments and demonstrates that, once the original point clouds are compressed into sparse trunk landmarks and compact topological structures, candidate screening, correspondence construction, and robust pose estimation can be efficiently performed using a small set of weighted vertex correspondences.

Overall, the real-world experiments further confirm the principal conclusions obtained from the simulated data. Repetitive trunk structures and local feature ambiguities substantially affect methods based on general geometric features, whereas trunk-level topological representations provide more stable structural constraints for forest submap registration. The proposed method achieves high registration accuracy across all three real-world plots while maintaining subsecond processing time, demonstrating a favorable accuracy--efficiency balance for real-world multi-submap forest mapping.

\subsection{Ablation Study and Back-End Comparison}

The preceding experiments demonstrate that the proposed method achieves a favorable accuracy--efficiency balance on both simulated and real-world datasets. To further investigate the sources of this performance, this section validates the underlying mechanisms from two perspectives. First, FGR, TEASER, and Quatro are evaluated as alternative back-end solvers under the same forest loop closure detection setting to examine the influence of different robust pose estimators on registration performance. Second, a unit-weight ablation is conducted to determine whether the reliability weights derived from strong and weak vertex support relationships improve the stability of back-end robust estimation.

\begin{figure*}[!t]
\centering
\includegraphics[width=0.92\textwidth]{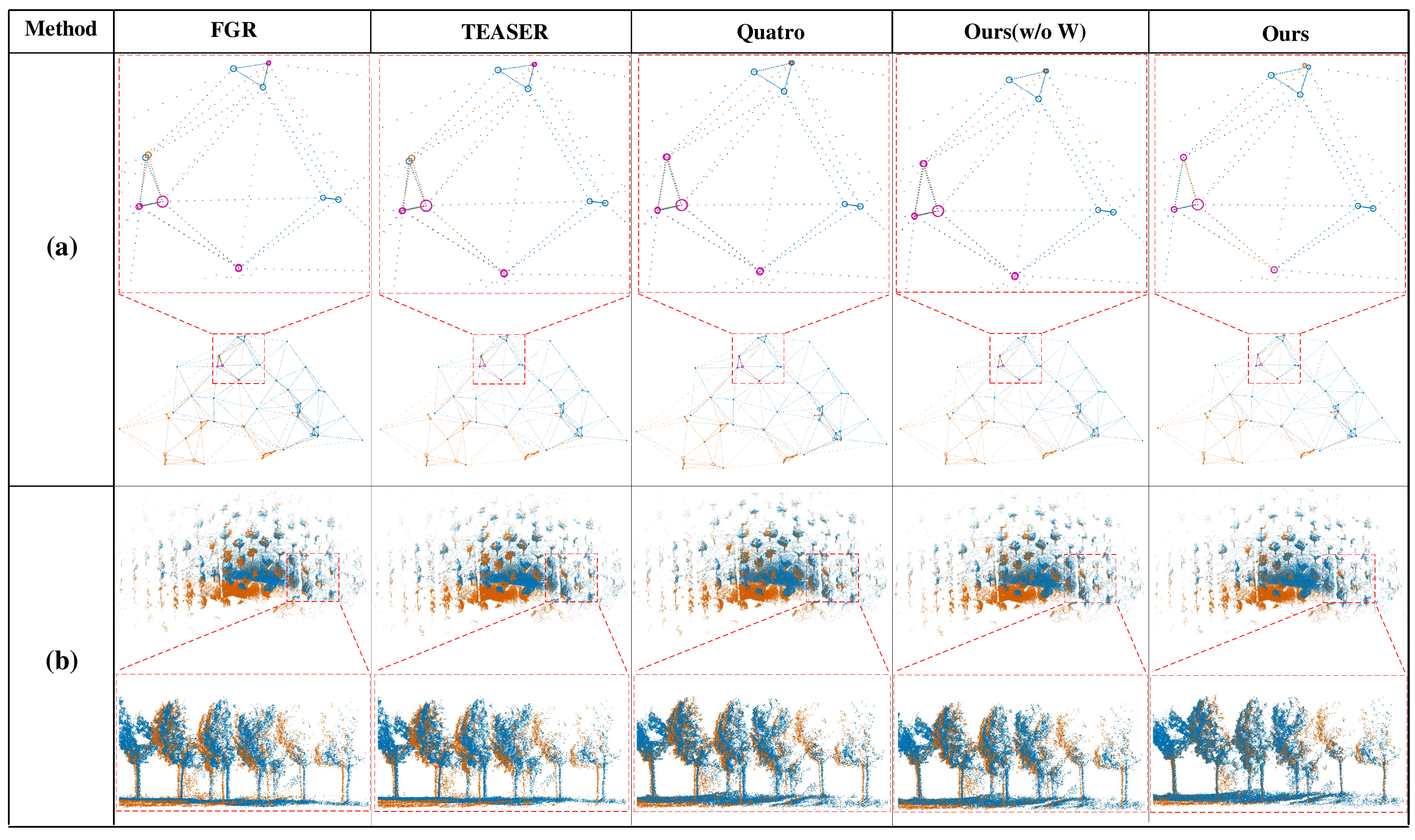}
\caption{Qualitative comparison of back-end solvers and ablation of topology-derived reliability weighting. Columns correspond to FGR, TEASER, Quatro, Ours (w/o W), and Ours. (a) Aligned query and candidate Delaunay topologies, shown in blue and orange, respectively; magenta circles indicate the final inlier vertices retained by each solver, and the red dashed boxes enlarge representative local topology regions. (b) Corresponding point cloud registration results, with blue and orange denoting the query and candidate submaps, respectively; the enlarged views show the local alignment of tree structures within the selected regions.}
\label{fig_6}
\end{figure*}

Fig.~\ref{fig_6} presents representative qualitative results of the back-end comparison and weighting ablation under the same front-end vertex correspondences. As shown in Fig.~\ref{fig_6}(a), the aligned Delaunay topologies obtained by different solvers exhibit clearly different local consistency patterns, and the magenta circles indicate the final inlier vertices retained by each method. FGR and TEASER produce evident topology misalignment, which is further reflected in the corresponding point cloud registration results in Fig.~\ref{fig_6}(b). These observations indicate that, in this representative case, the two general-purpose robust solvers are still affected by the large proportion of ambiguous vertex correspondences generated by repetitive forest structures.

Quatro and the unit-weight variant, denoted as Ours (w/o W), achieve substantially better alignment than FGR and TEASER and produce comparable qualitative results. Nevertheless, residual inconsistencies remain visible in both the enlarged topology regions and the local point cloud views. In contrast, the full method yields the closest overlap between the blue query submap and the orange candidate submap, together with a more consistent set of retained inlier vertices. The comparison between Ours (w/o W) and Ours shows that the improvement is mainly introduced by the topology-derived reliability weights, which increase the influence of repeatedly supported vertex correspondences while reducing the contribution of weakly supported or ambiguous matches during horizontal rotation and translation estimation. The corresponding quantitative analysis is presented below.

\subsubsection{Comparison of Back-End Solvers}

\begin{table*}[!t]
\caption{Back-End Comparison and Weighting Ablation Results on Simulated Data}
\label{tab:table3}
\centering
\scriptsize
\setlength{\tabcolsep}{2.5pt}

\begin{tabular}{|c|ccc|cc|cc|}
\hline

\multirow{2}{*}{Method}
& \multicolumn{3}{c|}{Average Pose Error}
& \multicolumn{2}{c|}{RMSE}
& \multicolumn{2}{c|}{Average Processing Time (ms)}
\\ \cline{2-8}

& Yaw (deg)
& Rot. Total (deg)
& Trans. Total (m)
& RMSE-R (deg)
& RMSE-T (m)
& Solver
& Total
\\ \hline

FGR
& 48.543 & 72.051 & 12.384
& 101.193 & 17.804
& 1.475 & 207.620
\\ \hline

TEASER
& 51.370 & 73.650 & 12.983
& 101.896 & 18.089
& 12.650 & 247.494
\\ \hline

Quatro
& 1.464 & 2.290 & 1.474
& 6.338 & 4.586
& 0.149 & 207.187
\\ \hline

Ours (w/o W)
& 3.137 & 3.982 & 1.272
& 15.824 & 4.148
& 1.673 & 217.391
\\ \hline

Ours
& 0.098 & 0.889 & 0.445
& 1.066 & 0.537
& 0.721 & 214.289
\\ \hline

\end{tabular}

\vspace{2pt}
\begin{minipage}{\textwidth}
\footnotesize
\textit{Note:} RMSE-R and RMSE-T denote the root mean square
errors of the total rotation and translation errors, respectively.
Solver and Total denote the average back-end solver time and average
total processing time, respectively.
\end{minipage}
\end{table*}

Table~\ref{tab:table3} presents the back-end comparison and weighting ablation results on the simulated dataset. FGR and TEASER both produce large yaw, total rotation, and total translation errors in the simulated forest environment. Although they are widely used robust global registration back ends, they are primarily designed to suppress outliers in general three-dimensional correspondences and do not explicitly exploit the horizontal structural characteristics or trunk-topology priors of forest point clouds. In environments containing repetitive trunk structures, the tentative correspondences generated by the front end may include numerous ambiguous matches with similar local geometric characteristics. When these correspondences are passed directly to general-purpose robust solvers, they can still result in incorrect horizontal rotation estimates and large translation errors. Compared with FGR and TEASER, Quatro substantially reduces the pose estimation errors. This result indicates that rotation--translation decoupling and quasi-$SO(3)$ rotation estimation are more suitable for the registration of gravity-aligned forest submaps. Mobile mapping data collected in forest environments generally provide reliable constraints along the gravity direction. Therefore, directly estimating the complete six-degree-of-freedom pose may not be the most effective strategy. Instead, concentrating the estimation on yaw and translation reduces the influence of degenerate degrees of freedom and incorrect correspondences on the estimated pose. Nevertheless, Quatro does not explicitly exploit the reliability weights constructed by the proposed front end, nor does it employ the XOY/H decoupling strategy specifically designed for gravity-aligned forest submaps. Consequently, although Quatro substantially outperforms FGR and TEASER on the simulated dataset, its total rotation and translation errors remain higher than those of the full method. The proposed method further reduces the yaw, total rotation, and total translation errors while maintaining a low total processing time. These results demonstrate the complementary benefits of reliability modeling for weighted vertex correspondences and decoupled robust pose estimation.

\begin{table*}[!t]
\caption{Back-End Comparison and Weighting Ablation Results on Real Data}
\label{tab:table4}
\centering
\scriptsize
\setlength{\tabcolsep}{2.5pt}

\begin{tabular}{|c|c|ccc|cc|cc|}
\hline

\multirow{2}{*}{Plot}
& \multirow{2}{*}{Method}
& \multicolumn{3}{c|}{Average Pose Error}
& \multicolumn{2}{c|}{RMSE}
& \multicolumn{2}{c|}{Average Processing Time (ms)}
\\ \cline{3-9}

&
& Yaw (deg)
& Rot. Total (deg)
& Trans. Total (m)
& RMSE-R (deg)
& RMSE-T (m)
& Solver
& Total
\\ \hline

\multirow{5}{*}{Real Plot 1}
& FGR
& 79.560 & 113.937 & 35.608
& 127.400 & 41.043
& 5.529 & 408.516
\\ \cline{2-9}

& TEASER
& 78.699 & 113.916 & 35.312
& 127.755 & 40.860
& 3.525 & 404.067
\\ \cline{2-9}

& Quatro
& 3.187 & 3.214 & 1.787
& 17.305 & 10.582
& 0.571 & 490.845
\\ \cline{2-9}

& Ours (w/o W)
& 4.170 & 4.191 & 1.827
& 20.920 & 8.130
& 0.513 & 416.017
\\ \cline{2-9}

& Ours
& 0.106 & 0.176 & 0.103
& 0.307 & 0.203
& 0.033 & 390.081
\\ \hline

\multirow{5}{*}{Real Plot 2}
& FGR
& 80.020 & 113.061 & 35.021
& 128.295 & 41.161
& 4.585 & 404.821
\\ \cline{2-9}

& TEASER
& 78.677 & 111.909 & 34.493
& 127.747 & 40.777
& 2.624 & 392.610
\\ \cline{2-9}

& Quatro
& 5.331 & 5.322 & 1.814
& 27.003 & 8.528
& 0.466 & 392.144
\\ \cline{2-9}

& Ours (w/o W)
& 4.621 & 4.622 & 1.168
& 26.283 & 5.860
& 0.478 & 393.324
\\ \cline{2-9}

& Ours
& 0.282 & 0.271 & 0.108
& 1.634 & 0.559
& 0.079 & 384.960
\\ \hline

\multirow{5}{*}{Real Plot 3}
& FGR
& 79.903 & 113.214 & 28.681
& 127.163 & 34.545
& 4.316 & 421.609
\\ \cline{2-9}

& TEASER
& 79.733 & 112.639 & 28.208
& 127.188 & 34.182
& 1.392 & 403.942
\\ \cline{2-9}

& Quatro
& 0.112 & 0.158 & 0.074
& 0.314 & 0.123
& 0.078 & 407.219
\\ \cline{2-9}

& Ours (w/o W)
& 4.335 & 4.386 & 1.244
& 22.068 & 5.760
& 0.098 & 415.640
\\ \cline{2-9}

& Ours
& 0.110 & 0.125 & 0.067
& 0.275 & 0.117
& 0.046 & 407.841
\\ \hline

\end{tabular}

\vspace{2pt}
\begin{minipage}{\textwidth}
\footnotesize
\textit{Note:} RMSE-R and RMSE-T denote the root mean square
errors of the total rotation and translation errors, respectively.
Solver and Total denote the average back-end solver time and average
total processing time, respectively.
\end{minipage}
\end{table*}

Table~\ref{tab:table4} further presents the back-end comparison and ablation results for the three real-world forest plots. The overall trend is consistent with that observed on the simulated dataset. FGR and TEASER continue to exhibit substantial yaw and total translation errors on the real-world forest data, indicating that general-purpose robust back ends remain susceptible to ambiguous correspondences under repetitive trunk structures and local observation differences. Quatro substantially outperforms FGR and TEASER across all three plots and achieves particularly high registration accuracy in Real Plot 3. This result further confirms the effectiveness of decoupled pose estimation for gravity-aligned forest point-cloud registration. Nevertheless, the performance of Quatro varies across the three plots. In particular, its total rotation and translation errors in Real Plot 1 and Real Plot 2 remain noticeably higher than those of the full method. This observation indicates that a general-purpose decoupled back end alone is insufficient to completely resolve structural ambiguities in forest environments. The proposed method consistently maintains lower yaw and total translation errors across the three real-world plots, demonstrating that the combination of front-end reliability weights and XOY/H-decoupled robust estimation improves pose estimation stability in real-world forest environments.

\subsubsection{Ablation Study on Reliability Weights}

To evaluate the contribution of the reliability weights, the full method was compared with a unit-weight configuration. The unit-weight configuration retains the same front-end procedures, including trunk landmark extraction, Delaunay topology construction, and vertex correspondence generation, but assigns an identical weight to every tentative correspondence during back-end pose estimation. This configuration isolates the contribution of the reliability weights derived from the strong and weak vertex support statistics to the final registration result.

As shown in Table~\ref{tab:table3}, removing the reliability weights results in substantially larger yaw, total rotation, and total translation errors for Ours (w/o W) on the simulated dataset. The increase in yaw error indicates that ambiguous correspondences interfere more directly with horizontal rotation estimation when all correspondences are assigned equal weights. In forest environments, different trunks may exhibit similar local topological structures. If weakly and strongly supported correspondences have equal influence during optimization, incorrect or unstable correspondences produced by locally repetitive structures may receive excessive influence in the back-end estimator, resulting in biased yaw and horizontal translation estimates. The full method assigns reliability weights to tentative correspondences according to the accumulated strong and weak vertex support. Vertex correspondences repeatedly supported by locally consistent topological relationships receive higher weights, whereas ambiguous correspondences supported by only a few or incomplete local structures have less influence. Therefore, the reliability weights are used not only for correspondence screening but also as prior information in back-end robust estimation. This encourages the optimization process to rely more strongly on structurally stable vertex correspondences with repeated topological support. Consequently, the full method simultaneously reduces the total rotation error, total translation error, and RMSE values on the simulated dataset.

The real-world results in Table~\ref{tab:table4} further validate this effect. Across all three real-world plots, Ours substantially reduces the yaw and total translation errors compared with Ours (w/o W). In Real Plot 1 and Real Plot 2, the unit-weight configuration still exhibits clear horizontal rotation and translation deviations, whereas the full method reduces these errors to relatively low levels. In Real Plot 3, both Quatro and the full method achieve high registration accuracy, while Ours maintains better or comparable total translation and RMSE results. These observations demonstrate that reliability weighting consistently suppresses ambiguous correspondences caused by locally repetitive structures in real-world forest environments.

Introducing the reliability weights does not substantially increase the computational cost. On both the simulated dataset and the three real-world plots, Ours and Ours (w/o W) require comparable total processing time, whereas the full method provides clear improvements in accuracy and stability. This result demonstrates that reliability modeling based on strong and weak vertex support statistics improves the quality of back-end robust estimation with only limited additional computational overhead. In summary, the back-end comparison demonstrates that the general-purpose robust solvers FGR and TEASER have difficulty handling ambiguous correspondences in repetitive forest structures. The decoupled estimation strategy of Quatro substantially improves registration performance but does not model the reliability of trunk-topology correspondences. The ablation results further demonstrate that the reliability weights in the proposed method are not merely heuristic corrections, but provide an important mechanism for propagating local topological support from the front end to back-end optimization. By combining reliability weights with XOY/H-decoupled robust pose estimation, the proposed method achieves more stable yaw and translation estimates on both simulated and real-world datasets.

\section{Conclusion} \label{section:5} 
This paper addresses loop closure detection and relative pose estimation among multiple independently acquired point-cloud submaps in GNSS-degraded forest understory environments. A lightweight loop closure detection and robust global registration method based on stable trunk topology was proposed. To address the sparsity and noise of low-cost LiDAR point clouds, the instability of local geometric features, and repetitive trunk structures in forest environments, trunk landmarks are first extracted from gravity-aligned local point-cloud submaps. A Delaunay triangle topology is then constructed from the trunk centers to form a compact structural representation of the forest scene. Based on this representation, candidate submaps are screened using submap-level distributions of trunk spacing and trunk radius. Edge-length and endpoint-radius consistency are subsequently used to accumulate strong and weak vertex support, from which reliability-weighted vertex correspondences are constructed. Finally, by exploiting the structural characteristics of gravity-aligned forest submaps, a reliability-weighted XOY/H-decoupled robust pose estimation strategy is developed. The recovery of the three-dimensional relative pose is decomposed into horizontal rotation, horizontal translation, and elevation translation estimation, improving registration robustness and computational efficiency under high outlier ratios.

Experiments on simulated data and three real-world forest plots demonstrate that baseline methods based on planar, corner, or general local geometric features are susceptible to repetitive trunk structures and local geometric ambiguities, particularly when estimating yaw and horizontal translation. In contrast, the proposed method exploits the spatial topology of tree trunks to provide more stable structural constraints and achieves high relative pose estimation accuracy on both simulated and real-world datasets. Compared with the high-accuracy robust registration method GROR, the proposed method achieves comparable or better mean pose accuracy in most scenarios while substantially reducing the average processing time per submap. These results demonstrate the strong potential of the proposed method for online forest loop closure detection and multi-submap relationship construction. The ablation study and back-end comparison further validate the effectiveness of the reliability weights and the XOY/H-decoupled robust estimator. Under the unit-weight configuration, weakly and strongly supported correspondences have equal influence during optimization, allowing ambiguous correspondences caused by repetitive trunk structures to interfere with horizontal rotation and translation estimation. In contrast, the proposed reliability weights increase the contribution of correspondences repeatedly supported by locally consistent topological relationships, thereby improving pose estimation stability. Compared with back-end variants based on FGR, TEASER, and Quatro, the proposed method further integrates front-end topological priors specific to forest environments with a decoupled estimation strategy tailored to gravity-aligned submaps, resulting in more stable registration performance under complex forest understory conditions.

One limitation is that elevation estimation remains dependent on the observation quality of trunk bases and near-ground points. Strong point-cloud downsampling, sparse ground returns, or occluded trunk bases can degrade trunk-base elevation estimation and increase the $z$-axis translation error. Future work will investigate more robust trunk-base estimation and terrain constraint modeling. Multi-submap joint optimization and cross-platform collaborative validation will also be explored to improve the long-term stability and generalization capability of the proposed method in larger-scale and more complex forest environments.


\bibliographystyle{IEEEtran}
\bibliography{DTIF}

\vfill

\end{document}